\documentclass{article} % For LaTeX2e
\usepackage{iclr2027_conference,times}
\iclrfinalcopy
\renewcommand{\headrulewidth}{0pt}

\usepackage{amsmath,amsfonts,bm}

\def\eqref#1{equation~\ref{#1}}
\def\1{\bm{1}}

\DeclareMathAlphabet{\mathsfit}{\encodingdefault}{\sfdefault}{m}{sl}
\SetMathAlphabet{\mathsfit}{bold}{\encodingdefault}{\sfdefault}{bx}{n}

\usepackage{hyperref}
\usepackage{url}

\title{Benchmarking and Enhancing Skill-Level Memory for Partially Observable Robotic Manipulation}

\author{
    \textbf{Yansong Shi}$^{1,2}$~
    \textbf{Jiange Yang}$^{5}$~
    \textbf{Xijie Yang}$^{3,2}$~
    \textbf{Shaowei Zhang}$^{4,2}$~
    \textbf{Yuhan Zhu}$^{5}$~
    \textbf{Tao Lu}$^{2}$~\\
    \textbf{Limin Wang}$^{5,2,}$\thanks{Corresponding author.}\\
  $^1$School of Information Science And Technology, University of Science and Technology of China \\
  $^2$Shanghai Artificial Intelligence Laboratory\\
  $^3$Computer Science and Technology, Zhejiang University \\
  $^4$Computer Science and Technology, Shanghai Jiaotong University \\
  $^5$State Key Lab of Novel Software Technology, Nanjing University
}

\usepackage{graphicx}
\usepackage{subcaption}
\usepackage{booktabs}
\usepackage{multirow}
\usepackage{graphicx}  % for \resizebox if needed
\usepackage{enumitem}
\usepackage[table]{xcolor}

\usepackage[T1]{fontenc}
\usepackage{booktabs}
\usepackage{tabularx}
\usepackage{array}
\usepackage{textcomp}
\usepackage{wrapfig}

\usepackage{makecell}
\usepackage{booktabs}
\usepackage[capitalise]{cleveref}
\crefname{figure}{Figure}{Figures}
\crefname{table}{Table}{Tables}
\crefname{section}{Section}{Sections}
\crefname{equation}{Equation}{Equations}

\usepackage{hyperref}

\usepackage{booktabs}
\usepackage{array}
\usepackage{tabularx}
\usepackage{adjustbox}
\usepackage{collcell}

\newcommand{\HIDElang}[1]{%
  \adjustbox{max width=\linewidth}{\scriptsize #1}%
}

\newcolumntype{Q}{>{\collectcell\HIDElang}X<{\endcollectcell}}
\usepackage{xcolor}
\usepackage{tcolorbox}
\tcbuselibrary{skins,breakable}
\usepackage{needspace}

\definecolor{HIDEBlue}{HTML}{5484B0}
\definecolor{HIDEGreen}{HTML}{588C78}
\definecolor{HIDEPurple}{HTML}{8D75A5}
\definecolor{HIDEOrange}{HTML}{D99A5E}
\definecolor{HIDEGray}{HTML}{666666}
\definecolor{darkGreen}{RGB}{92, 148, 110}

\newcommand{\HIDEcategory}[3]{%
  \par\addvspace{10pt}%
  \Needspace{18\baselineskip}%
  \begin{tcolorbox}[
    enhanced,
    frame hidden,
    colback=#1!12!white,
    borderline west={2pt}{0pt}{#1!75!black},
    arc=1.5mm,
    boxsep=0pt,
    left=9pt,
    right=9pt,
    top=6pt,
    bottom=6pt,
    before skip=0pt,
    after skip=6pt
  ]
    {\bfseries #2}\hfill{}\par
    {\small #3}
  \end{tcolorbox}%
}

\newtcolorbox{HIDEtask}[2]{
  enhanced,
  breakable=false,
  colback=#1!2!white,
  colframe=#1!25!white,
  colbacktitle=#1!10!white,
  coltitle=black,
  title={#2},
  title after break={#2\space (continued)},
  fonttitle=\bfseries\small,
  fontupper=\small,
  boxrule=0.35pt,
  arc=1.5mm,
  boxsep=0pt,
  left=8pt,
  right=8pt,
  top=5pt,
  bottom=6pt,
  toptitle=4pt,
  bottomtitle=4pt,
  before skip=7pt,
  after skip=7pt,
  before upper={%
    \setlength{\parindent}{0pt}%
    \setlength{\parskip}{2.5pt}%
  }
}

\newcommand{\HIDEfield}[2]{%
  \textbf{#1:} #2\par
}

\newcommand{\HIDEstats}[2]{%
  \textbf{Variation Number:} #1\qquad
  \textbf{Keyframes:} #2\par
}

\newcommand{\HIDEslot}[1]{%
  \textnormal{\texttt{\textless#1\textgreater}}%
}

\newcommand{\HIDEinstruction}[1]{%
  {\raggedright
   \textbf{Language Instructions:} \textit{``#1''}\par}%
}

\begin{document}

\maketitle
% %%%%%%%%%%%%%%%%%%%%%%%%%%%%%%%%%%% arxiv
% \setlength{\droptitle}{-20pt}

\vspace{-10pt}
\begin{abstract}
Recent advances in robot learning have enabled manipulation policies to perform increasingly diverse tasks and generalize across environments. However, reliable execution often depends on hidden task states that cannot be determined from current observations alone, making interaction history essential. We introduce \textbf{HIDE}, a benchmark for evaluating manipulation memory under partial observability. HIDE comprises 15 tasks covering repetition counting, historical-state recall, and execution-progress tracking, with randomized initial configurations and decision points where similar observations require different actions depending on prior events. We further propose \textbf{SEEK}, a framework combining three complementary memory mechanisms to retain historical evidence and track execution state. Evaluations reveal substantial limitations in existing policies on HIDE, while memory augmentation improves task success in both simulation and real-world experiments. Individual mechanisms benefit some tasks but can degrade others; their combination achieves the highest average success rate on HIDE among the evaluated configurations. These findings highlight the importance of maintaining internal representations of hidden task states and matching memory design to task-specific information requirements.
Project page: \href{https://nanamma.github.io/HIDE-SEEK/}{HIDE-SEEK}.
\end{abstract}

\section{Introduction}

Visual robotic manipulation is rapidly progressing toward general-purpose
control. Recent robot foundation models, including
$\pi_0$~\citep{black2024pi0}, OpenVLA~\citep{kim2024openvla}, and
GR00T N1~\citep{bjorck2025gr00t}, leverage large-scale robot data for
diverse manipulation, while world-action models explore predictive action
generation through future environment modeling~\citep{ye2026gigaworld,wang2026wam}.
However, most manipulation policies still rely primarily on current
observations or limited temporal context. In real-world execution,
critical decision-relevant information may be hidden from the current
observation and must instead be inferred from interaction history.

Recent systems separate high-level reasoning from low-level execution
through language subgoals~\citep{shi2025hirobot,physicalintelligence2025pi05}
or learned semantic interfaces~\citep{figure2025helix,bjorck2025gr00t}.
However, specifying a task objective does not necessarily provide the
execution state required by a manipulation policy. For example, an
instruction to repeat an operation twice does not indicate how many
executions have already been completed. Similarly, object references and
task progress may become unavailable during execution. Therefore, policies
must maintain relevant internal states rather than relying only on
externally specified goals.

We define these decision-relevant variables as \emph{hidden task states}:
information that cannot be inferred from the current visual and
proprioceptive observations alone but can be recovered from interaction
history. Different histories may produce visually similar observations
while requiring different actions, corresponding to observation aliasing
in a POMDP~\citep{kaelbling1998planning}. Such states include completed
repetitions, historical references, and procedural progress.

To study this problem, we introduce \textbf{HIDE}, a benchmark for
hidden-state memory in robotic manipulation. HIDE contains 15 tasks
across three categories: \emph{repetition counting},
\emph{historical-state recall}, and \emph{execution-progress tracking}.
Built on RLBench~\citep{james2020rlbench}, it provides randomized
configurations and appearance variations, automated demonstration
generation, and structured hidden-state annotations for memory analysis.

Existing memory-based policies retain history through recurrent states,
memory banks, or retrieval mechanisms. Methods such as
SAM2Act+~\citep{fang2025sam2act}, MemoryVLA~\citep{shi2025memoryvla}, and
Embodied-SlotSSM~\citep{chung2026liberomem} demonstrate the benefits of
historical representations. However, retaining history does not ensure
recovery of the required hidden states. RoboMME~\citep{dai2026robomme}
further shows task-dependent memory effectiveness, motivating memory
designs tailored to hidden-state requirements.

To address this challenge, we introduce \textbf{SEEK}, a memory-augmented
manipulation framework that maintains recent interaction context,
persistent historical references, and execution progress. Evaluations on
HIDE reveal substantial limitations of existing policies, while SEEK
improves performance in both simulation and real-world experiments.
Individual mechanisms exhibit distinct capability profiles, benefiting
some hidden-state requirements while potentially degrading others.
Their combination achieves the strongest overall performance.

\begin{figure*}[t!]
    \vspace{-40pt}
    \centering
    \includegraphics[width=0.95\linewidth]{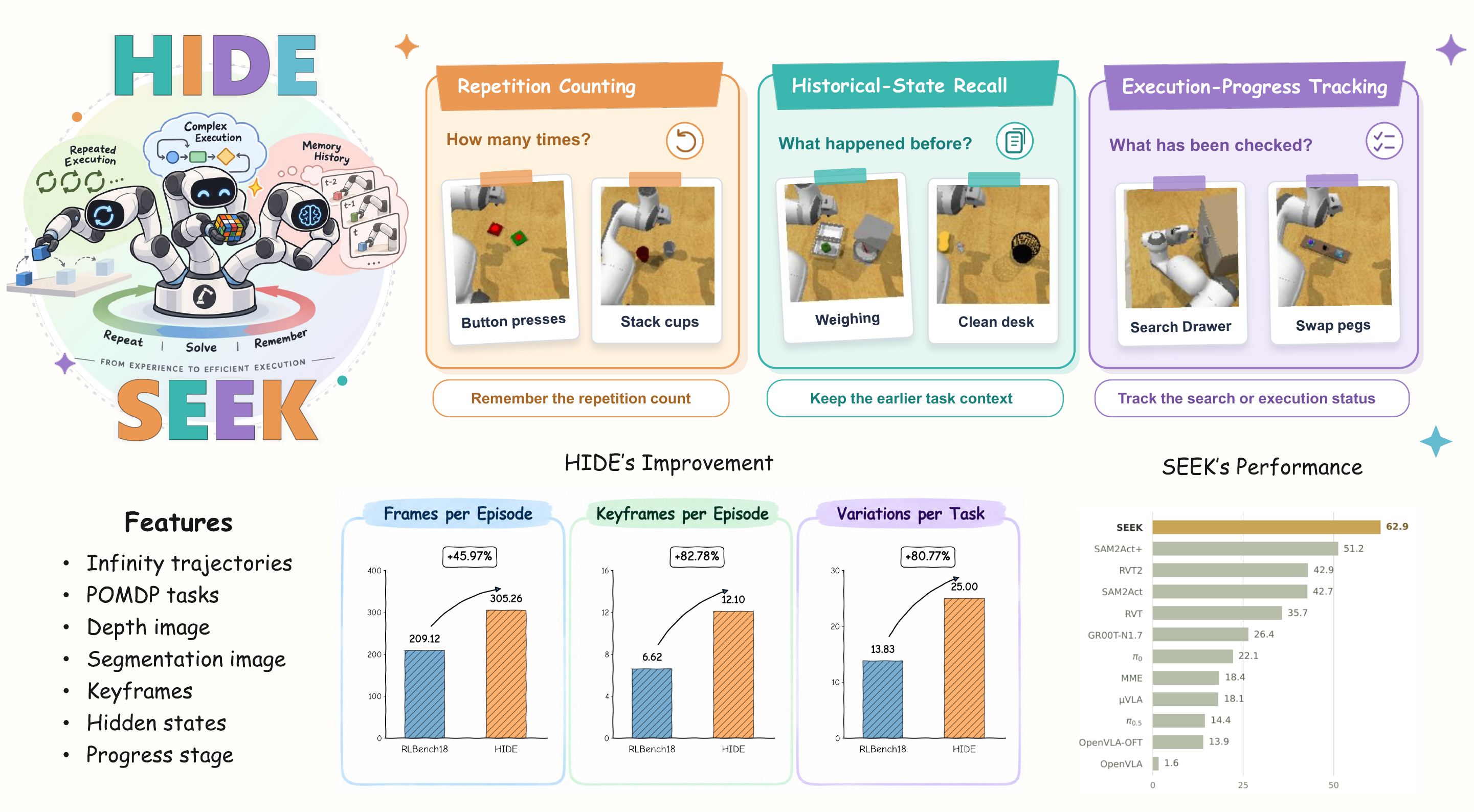}
    \vspace{-10pt}
    \caption{
        \textbf{Overview of HIDE and SEEK.}
        HIDE covers three hidden-state requirements and provides longer,
        more diverse trajectories than RLBench, while SEEK achieves the
        highest overall performance on the benchmark.
    }
    \label{fig:main}
    \vspace{-10pt}
\end{figure*}

Our contributions are threefold:
\begin{itemize}[leftmargin=1.2em, labelsep=0.4em,
               itemsep=2pt, topsep=2pt, parsep=0pt]

\item \textbf{Hidden-state benchmark.}
We formulate manipulation memory from decision-relevant hidden states and
introduce HIDE, a 15-task benchmark covering repetition counting,
historical-state recall, and execution-progress tracking.

\item \textbf{Memory framework and analysis.}
We introduce SEEK with three complementary memory mechanisms and
systematically analyze their individual and combined effects through
ablations and cross-category evaluation.

\item \textbf{Performance gains and insights.}
SEEK improves manipulation performance in simulation and real-world
experiments, while HIDE reveals how different memory mechanisms correspond
to distinct hidden-state requirements.
\end{itemize}

\section{Related Work}

\subsection{Memory-Dependent Robotic Benchmarks}

RLBench~\citep{james2020rlbench},
CALVIN~\citep{mees2022calvin},
LIBERO~\citep{liu2023libero}, and
RoboCasa~\citep{nasiriany2024robocasa} provide diverse manipulation
tasks for evaluating generalization and long-horizon execution.
However, these benchmarks primarily focus on task completion and
generalization rather than explicitly isolating the historical
information required for decision making. Memory-oriented benchmarks
such as MemoryBench~\citep{fang2025sam2act} and
MIKASA-Robo~\citep{cherepanov2025memory} investigate
memory-dependent behaviors under partial observability. Recent
benchmarks explore aspects of manipulation memory:
RoboMME~\citep{dai2026robomme} organizes tasks around temporal,
spatial, object, and procedural memory; RoboMemArena~\citep{lei2026robomemarena}
studies counting, occlusion, object transfer, and sequential execution;
RMBench~\citep{chen2026rmbench} characterizes memory complexity through
decision-critical historical observations; and LIBERO-Mem~\citep{chung2026liberomem}
focuses on object-level interaction histories.

HIDE complements these benchmarks by organizing tasks according to the
hidden variables required for correct manipulation decisions, including
repetition count, historical scene state, and execution progress. Instead
of measuring memory demand only through task length, scene complexity,
or the amount of retained context, HIDE explicitly constructs decision
points where the current observation is insufficient but interaction
history provides the necessary evidence. Built upon RLBench, HIDE
supports configurable initializations, appearance variations, and
task-specific execution conditions through automated demonstration
collection, enabling controlled evaluation of different hidden-state
requirements and memory mechanisms.

\subsection{Memory-Augmented Robotic Policies}

Existing manipulation policies incorporate historical information through
temporal windows, recurrent states, or explicit memory representations.
HistRISE~\citep{chen2026history} models object dynamics using point
trajectories, ContextVLA~\citep{jang2025contextvla} compresses multi-frame
visual context, MemoryVLA~\citep{shi2025memoryvla} maintains perceptual
and cognitive memories, and SAM2Act+~\citep{fang2025sam2act} integrates
a visual memory bank. Retrieval-based approaches such as MemER
~\citep{sridhar2026memer} and PrediMem in RoboMemArena
~\citep{lei2026robomemarena} select relevant historical information for
long-horizon control. These methods demonstrate that historical
representations can improve manipulation, while also introducing
trade-offs between temporal coverage, information retention, and
online memory maintenance.

However, retaining historical observations alone does not guarantee that
a policy can recover the hidden state required for a specific decision.
Recent studies also investigate structured memory combinations:
Mem-0 in RMBench~\citep{chen2026rmbench} combines anchor and sliding
memories with subtask termination prediction, while RoboMME
~\citep{dai2026robomme} analyzes the effect of different memory
representations and integration strategies. Different from these works,
we study memory according to explicit hidden-state requirements in
manipulation execution. SEEK introduces complementary mechanisms for
recent context, persistent historical references, and execution progress,
and evaluates their individual benefits, limitations, and interactions
through controlled cross-category experiments.
\section{HIDE}
\label{sec:benchmark}
\subsection{Hidden-State-Dependent Manipulation} %%%%%%%%%%%%%%%%%%%%%%%%%%%%%%%%%%%%%%%

We formulate robotic manipulation as a partially observable decision
process. At timestep $t$, the agent receives a task instruction $g$,
current observation $o_t$, and optionally its history
$h_t=(o_{1:t},a_{1:t-1})$. The underlying state is decomposed as
$s_t=(x_t,z_t)$, where $x_t$ is the observable physical state and $z_t$
is a task-relevant latent state not identifiable from the current
observation alone.

The latent state summarizes the historical information required for
optimal decision making:
\begin{equation}
    \pi^*(a_t \mid h_t,g)
    =
    \pi^*(a_t \mid o_t,z_t,g).
\end{equation}
We define a task as \emph{hidden-state-dependent} if there exist two
interaction histories $h_t$ and $h'_t$ that lead to visually equivalent
current observations but require different optimal actions:
\begin{equation}
    \begin{aligned}
        d(o_t,o'_t) &\leq \epsilon, \\
        z_t &\neq z'_t, \\
        \pi^*(\cdot\mid h_t,g)
        &\neq
        \pi^*(\cdot\mid h'_t,g).
    \end{aligned}
\end{equation}
Thus, long horizons or visual occlusion alone do not make a task
memory-dependent; historical information must causally determine the
correct subsequent behavior.

\begin{figure*}[]
    \vspace{-20pt}
    \centering
    \includegraphics[width=0.99\linewidth]{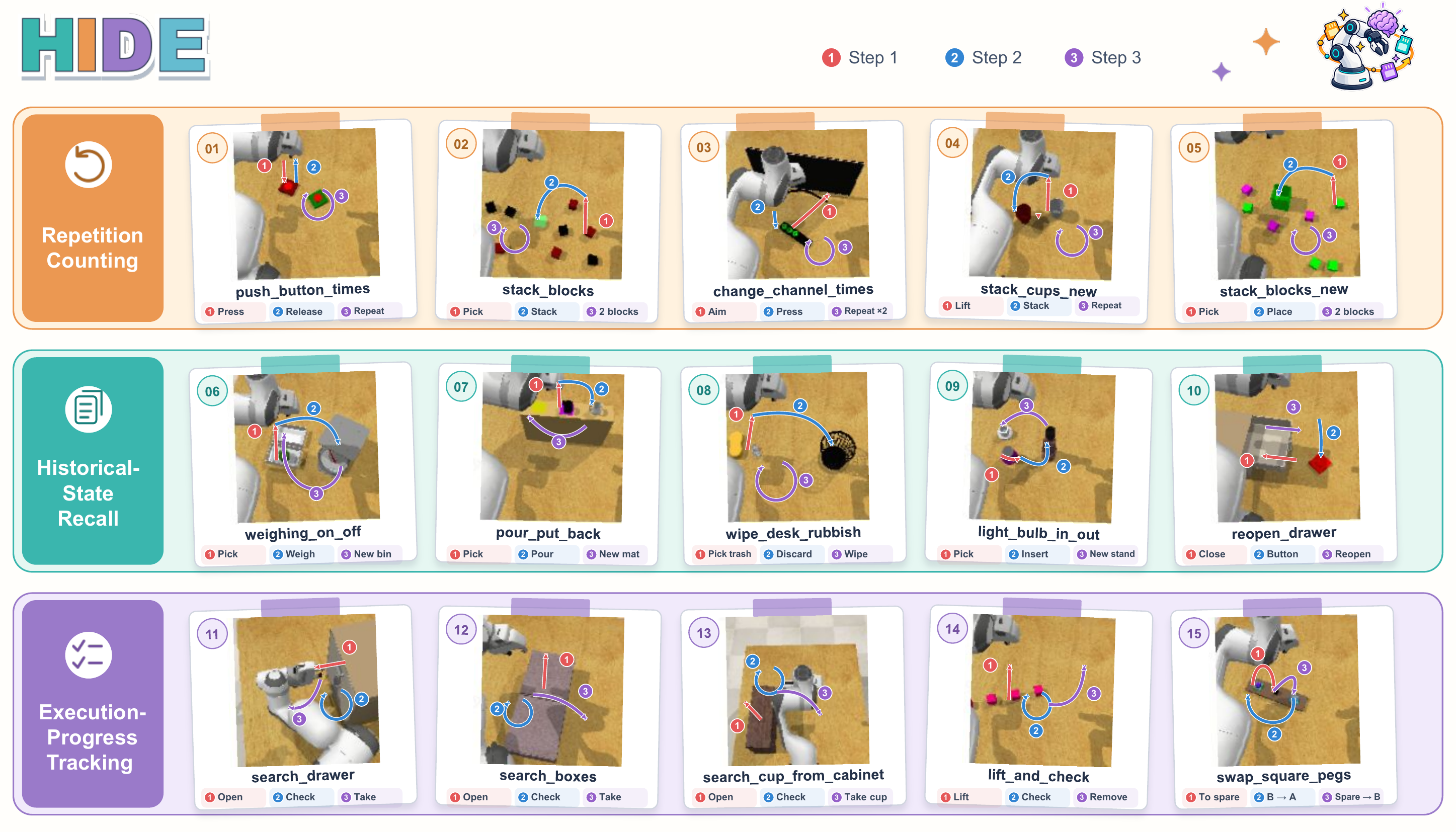}
    \caption{
        Detailed trajectories for the three HIDE task categories, with step
        order and repeated/checking operations indicated by colored arrows and labels.
    }
    \label{fig:tasks}
    \vspace{-10pt}
\end{figure*}
\subsection{Task Categories and Design}

HIDE organizes manipulation tasks into three categories according to their primary memory
requirements: repetition counting, historical-state recall, and execution-progress tracking.
Each category introduces decision points where the correct behavior depends on information
that cannot be recovered from the current observation alone.

\textbf{Repetition Counting (RS).} The robot must repeat an operation a specified number of times.
Since different repetitions can produce visually similar observations, the completed count is
ambiguous from the current scene. The robot must therefore track completed repetitions and
decide when to continue, stop, or switch. This category evaluates whether a policy can maintain
an accurate count throughout repetitive interactions rather than merely reproduce a recurring
action pattern.

\textbf{Historical-State Recall (HSR).} The robot must act on previously observed information, such
as an object's color, identity, location, or an earlier scene configuration. At the relevant
decision point, this information is no longer directly accessible due to occlusion, scene
changes, or visually confusable alternatives. This category evaluates history-conditioned
decisions rather than recognition from the current scene alone.

\textbf{Execution-Progress Tracking (EPT).} These tasks involve multiple manipulation substeps whose
intermediate observations may not uniquely reveal which steps have been completed. The robot
must use its execution history to determine current progress and select the appropriate next
action, avoiding unnecessary repetition, skipped steps, or premature termination. The emphasis
is on tracking progress under observational ambiguity rather than task length alone.

Tasks are grouped by their primary memory requirement, though individual tasks may involve
additional demands. Across all categories, occlusion, visual similarity, and repeated
interactions serve to create history dependence rather than as separate task categories.
Detailed task specifications are provided in
Figure~\ref{fig:tasks} and Appendix~\ref{app:sim_tasks}.

\subsection{Benchmark Construction} %%%%%%%%%%%%%%%%%%%%%%%%%%%%%%%%%%%%%%%%

We build HIDE upon RLBench~\citep{james2020rlbench}, which provides a flexible simulation framework with reusable robot, object, and scene assets. Each task is extended from the standard RLBench task-generation pipeline by specifying object initialization ranges, randomized scene configurations, and a sequence of manipulation keypoints. During demonstration generation, the simulator samples object poses within predefined regions and instantiates diverse visual appearances, including colors and materials. The robot then executes a continuous trajectory by following the corresponding manipulation keypoints, enabling a large number of task variations to be generated from the same task template.

Compared with conventional RLBench tasks, HIDE introduces longer and more memory-dependent manipulation sequences. We modify existing assets and interaction patterns to create repeated operations, historically dependent object choices, and multi-stage procedures whose correct execution cannot always be determined from the current observation alone. Task difficulty can be systematically varied through factors such as the number of repetitions, the number and similarity of distractor objects, the duration between informative observations and subsequent decisions, and the number of manipulation stages.

Demonstrations are collected using the standard RLBench scripted expert pipeline. After data collection, we automatically segment each trajectory into task stages according to the predefined manipulation keypoints and task-specific success conditions. These stage annotations provide execution-progress information for both training and evaluation, while avoiding additional manual annotation.

\section{SEEK}
\label{sec:method}

\begin{figure*}[t]
    \vspace{-10pt}
    \centering
    \includegraphics[width=0.95\linewidth]{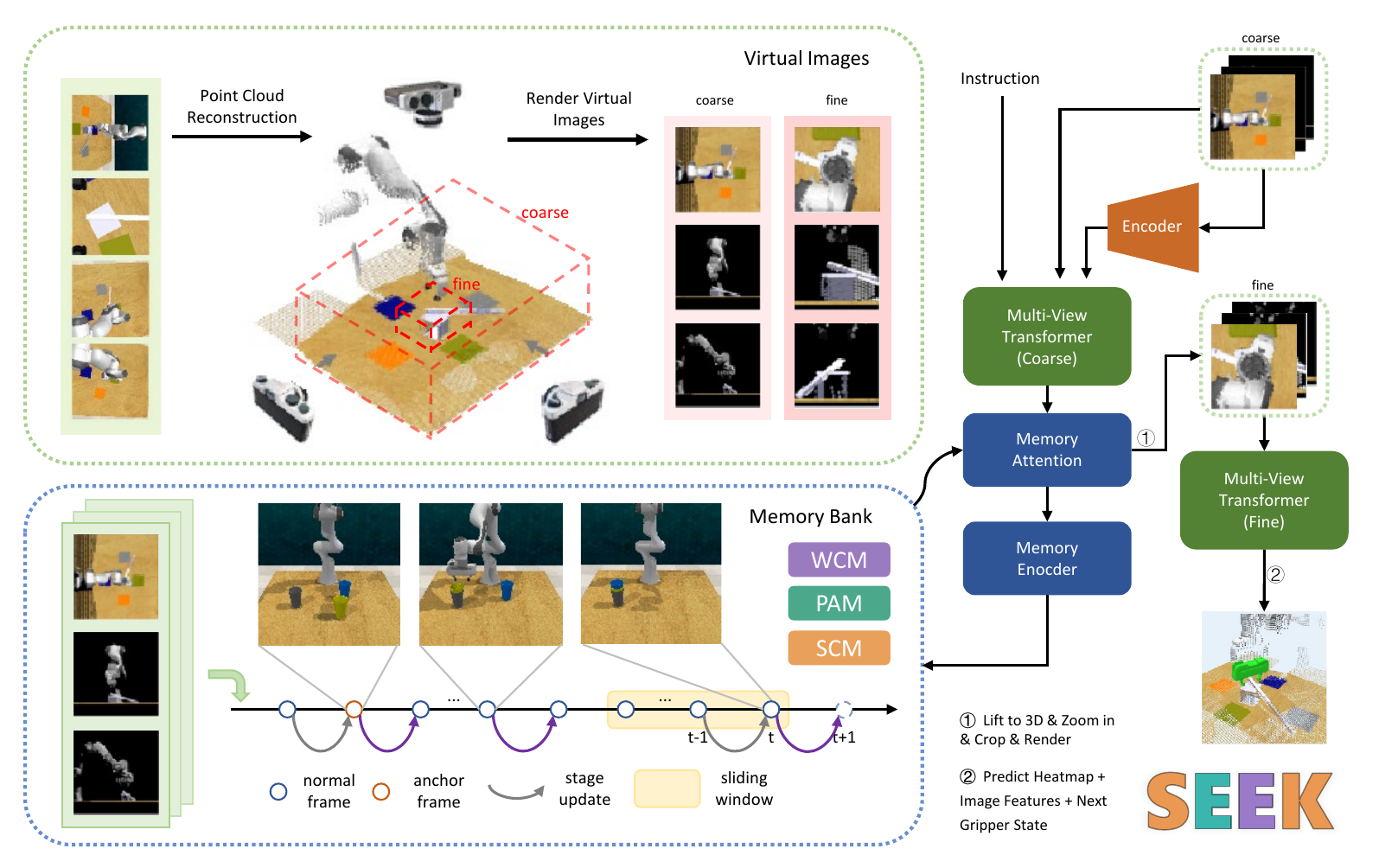}
    \caption{
        Overview of SEEK. Windowed Context Memory (WCM) retains recent interactions, Persistent Anchor Memory (PAM) retrieves relevant historical evidence, and Stage-Counter Memory (SCM) tracks execution progress. Together, they condition the coarse-to-fine manipulation policy.
    }
    \label{fig:method}
    \vspace{-10pt}
\end{figure*}

Under partial observability, a manipulation policy must distinguish states
that look similar but require different actions because of their history.
Recent interactions reveal local changes, earlier observations may contain
now-hidden evidence, and execution progress determines whether an action
should be repeated or the policy should move on.
SEEK organizes these dependencies into three complementary components:
\textbf{Windowed Context Memory (WCM)}, \textbf{Persistent Anchor Memory
(PAM)}, and \textbf{Stage-Counter Memory (SCM)}.
Together, they provide recent context, a selectively retrieved long-range
reference, and an explicit progress state within a compact attention input.

\subsection{Hidden-State Memory Modeling}
\label{sec:memory_modeling}

\paragraph{Windowed Context Memory.}
WCM retains the latest $K$ interaction memories in temporal order.
Each new entry replaces the oldest once the window is full, preserving
recent object changes and manipulation outcomes.
This local context helps the policy track what has just happened, but
cannot retain evidence indefinitely as an episode unfolds.
In sequential exploration, it can indicate which locations were recently
inspected and how their contents changed, providing context for the next
interaction without requiring the entire history at every decision.

\paragraph{Persistent Anchor Memory.}
PAM preserves access to evidence after it leaves the recent window.
For example, an early observation may reveal an object's identity or
location before later interactions occlude it.
Rather than using a fixed first-frame reference, PAM retrieves a
historical entry based on the current observation.
Let $\mathcal{I}_t^{W}$ denote the indices in WCM. The eligible archive is
$\mathcal{A}_t=\{i:0\leq i<t,\ i\notin\mathcal{I}_t^{W}\}$, and retrieval is
\begin{equation}
    a_t=\underset{i\in\mathcal{A}_t}{\operatorname{argmax}}
    \;\operatorname{cos}(\mathbf{k}_t,\mathbf{k}_i),
    \label{eq:anchor_retrieval}
\end{equation}
where $\mathbf{k}_t$ and $\mathbf{k}_i$ are pooled visual descriptors
computed before memory fusion.
Each descriptor is stored with its interaction memory, separating the
compact retrieval key from the spatial features used for action prediction.
After excluding entries already covered by WCM, PAM selects the most similar
reference from the remaining history.
The anchor is recomputed at each decision and omitted when no eligible entry
exists. Entries leaving WCM remain in the episode archive and become eligible
for PAM, preserving older evidence without duplicating recent context.

\paragraph{Stage-Counter Memory.}
SCM represents progress using a discrete counter and a learned stage
embedding. The counter starts at zero and advances when the policy predicts
a stage boundary.
Whereas WCM and PAM describe previous interactions, SCM indicates how far
the execution has progressed. This distinction is useful when repeated
actions return the scene to a similar appearance: visual similarity alone
may not distinguish an intermediate repetition from the final one.
The counter changes at predicted interaction boundaries rather than at every
control step, allowing progress to remain stable while an action unfolds.

\subsection{Memory Encoding and Retrieval}
\label{sec:memory_retrieval}
Multi-camera RGB-D observations are rendered into three virtual views.
For each view $v$, a shared memory encoder combines visual features
$\mathbf{X}_t^v$ with the predicted coarse translation heatmap
$\mathbf{H}_t^v$ to produce interaction memory $\mathbf{m}_t^v$,
associating scene content with the predicted manipulation target.
The memory is written only after action prediction and is available to
subsequent decisions. WCM and PAM select from these encoded entries,
while SCM provides a separate learned representation $\mathbf{S}_t$.
Retaining spatial feature maps preserves the location of interaction
evidence, allowing retrieval to provide localized rather than global
historical information.

The three memory components are combined as
\begin{equation}
    \mathcal{M}_t^v=
    [\mathcal{W}_t^v;\mathbf{m}_{a_t}^v;\mathbf{S}_t],
    \qquad
    \widetilde{\mathbf{X}}_t^v=
    \operatorname{MemoryAttn}(\mathbf{X}_t^v,\mathcal{M}_t^v).
    \label{eq:memory_attention}
\end{equation}
The anchor is omitted when no eligible archive entry exists. Spatial and
temporal encodings distinguish view structure and memory slots, while
SCM uses a separate position embedding. Following SAM~2~\citep{ravi2024sam},
current visual features attend to memory to retrieve task-relevant history
and progress. The read budget is bounded by at most $K$ recent entries, one
anchor, and one stage representation per view, regardless of archive length.

\subsection{Memory-Augmented Manipulation Policy}
\label{sec:policy}

SEEK builds on a language-conditioned, coarse-to-fine multi-view policy.
The coarse branch uses memory-conditioned features to predict workspace
translation heatmaps, whose target defines a local region for fine action
prediction, including position, rotation, and gripper state. Memory thus
influences target selection before local refinement, while the fine branch
does not independently retrieve the episode archive. Historical context
can disambiguate visually similar candidate objects, while the fine branch
refines the selected interaction using current local geometry. After
prediction, the interaction is stored for future WCM and PAM reads, and
stage-transition prediction updates SCM. All memories and the counter are
reset between episodes.

\subsection{Training Objective}
\label{sec:training_objective}

SEEK is trained by behavior cloning on expert demonstrations.
The action loss supervises coarse and fine translation, rotation, gripper
state, and collision-related predictions. A stage-boundary loss trains the
progress predictor, giving
$\mathcal{L}=\mathcal{L}_{\mathrm{act}}+
\lambda_{\mathrm{stage}}\mathcal{L}_{\mathrm{stage}}$.
Training sequences are processed in temporal order with annotated stage
states; inference updates memory and the counter online using the policy's
predictions. Only preceding observations are eligible for retrieval.

\section{Experiments}
\label{sec:experiments}

We organize our experiments around five research questions:

\begin{itemize}[leftmargin=0.3cm,labelsep=0.2cm,
                itemsep=2pt,parsep=0pt,topsep=0pt]
    \item \textbf{Q1:}
    What limitations do existing policies exhibit
    across HIDE's hidden-state requirements?

    \item \textbf{Q2:}
    How much does SEEK improve performance,
    and are gains consistent across categories?

    \item \textbf{Q3:}
    How do memory content and components
    affect overall and category-specific performance?

    % \item \textbf{Q4:}
    % How does performance vary with increasing repetition count,
    % temporal delay, and search complexity?

    \item \textbf{Q4:}
    Does the method remain effective on standard tasks,
    under perturbations, and in real-world execution?
\end{itemize}

%%%%%%%%%%%%%%%%%%%%%%%%%%%%%%%%%%%%%%%%%%%%%%%%%%%%%%%%%%%%%%%%%%%%%% HIDE main_results
\definecolor{HIDEOverall}{HTML}{8C8278}
\begin{table*}[t]
    \centering
    \caption{
        \textbf{Performance on HIDE.}
        Success rates (\%) on individual tasks and their category averages.
        The benchmark contains three categories:
        repetition counting, historical-state recall, and
        execution-progress tracking.
        The best and second-best results in each column are
        \textbf{bold} and \underline{underlined}, respectively,
        including ties.
    }
    \label{tab:main_results}

    \scriptsize
    \setlength{\tabcolsep}{2.4pt}
    \renewcommand{\arraystretch}{1.10}

    \resizebox{\textwidth}{!}{%
    \begin{tabular}{@{}l>{\columncolor{HIDEOverall!10}}c
                      ccccc >{\columncolor{HIDEOrange!12}}c
                      ccccc >{\columncolor{HIDEGreen!12}}c
                      ccccc >{\columncolor{HIDEPurple!12}}c@{}}
        \toprule

        &
        \cellcolor{white}
        & \multicolumn{6}{>{\columncolor{HIDEOrange!28}}c}
          {\textbf{Repetition Counting (RC)}}
        & \multicolumn{6}{>{\columncolor{HIDEGreen!28}}c}
          {\textbf{Historical-State Recall (HSR)}}
        & \multicolumn{6}{>{\columncolor{HIDEPurple!28}}c}
          {\textbf{Execution-Progress Tracking (EPT)}}
        \\
        \cmidrule(lr){3-8}
        \cmidrule(lr){9-14}
        \cmidrule(lr){15-20}

        \textbf{Method}
        & \textbf{Avg.}

        & \shortstack{Push\\Button}
        & \shortstack{Stack\\Blocks}
        & \shortstack{Change\\Channel}
        & \shortstack{Stack\\Cups}
        & \shortstack{Stack\\Blocks-N}
        & \textbf{Avg.}

        & \shortstack{Weigh\\On/Off}
        & \shortstack{Pour\\Back}
        & \shortstack{Wipe\\Desk}
        & \shortstack{Light\\Bulb}
        & \shortstack{Reopen\\Drawer}
        & \textbf{Avg.}

        & \shortstack{Search\\Drawer}
        & \shortstack{Search\\Boxes}
        & \shortstack{Search\\Cabinet}
        & \shortstack{Lift \&\\Check}
        & \shortstack{Swap\\Pegs}
        & \textbf{Avg.}
        \\

        \midrule

        OpenVLA~\citep{kim2024openvla}
        & 1.6
        & 8 & 0 & 4 & 0 & 4 & 3.2
        & 8 & 0 & 0 & 0 & 0 & 1.6
        & 0 & 0 & 0 & 0 & 0 & 0.0
        \\

        OpenVLA-OFT~\cite{kim2025fine}
        & 13.9
        & 12 & 0 & 28 & 0 & 0 & 8.0
        & 36 & 0 & 68 & 0 & 32 & 27.2
        & 0 & 4 & 0 & 28 & 0 & 6.4
        \\

        $\pi_0$~\citep{black2024pi0}
        & 22.1
        & 40 & 0 & 12 & 0 & 0 & 10.4
        & 40 & 24 & 92 & 0 & 24 & 36.0
        & 12 & \underline{28} & 36 & 24 & 0 & 20.0
        \\

        $\pi_{0.5}$~\citep{physicalintelligence2025pi05}
        & 14.4
        & 16 & 0 & 4 & 4 & 4 & 5.6
        & 20 & 8 & 76 & 0 & 20 & 24.8
        & 4 & 8 & 20 & 32 & 0 & 12.8
        \\

        GR00T-N1.7~\citep{bjorck2025gr00t}
        & 26.4
        & 12 & 0 & 12 & 4 & 4 & 6.4
        & \underline{96} & 0 & 84 & 0 & 68 & \underline{49.6}
        & 28 & 16 & 40 & 24 & 8 & 23.2
        \\

        RVT~\citep{goyal2023rvt}
        & 35.7
        & \underline{68} & 12 & 20 & \underline{40} & 48 & 37.6
        & 28 & 24 & 80 & 0 & 60 & 38.4
        & 40 & 12 & 44 & 60 & 0 & 31.2
        \\

        RVT2~\citep{goyal2024rvt}
        & 42.9
        & 12 & \textbf{56} & 24 & \textbf{48} & \underline{76}
        & 43.2
        & 36 & \textbf{36} & 84 & 8 & 48 & 42.4
        & 48 & 16 & 56 & \underline{88} & 8 & 43.2
        \\

        SAM2Act~\citep{fang2025sam2act}
        & 42.7
        & 16 & \textbf{56} & 12 & \textbf{48} & \underline{76}
        & 41.6
        & 8 & \underline{32} & \textbf{100} & 4 & 52 & 39.2
        & 48 & 24 & 64 & 84 & 16 & 47.2
        \\

        \midrule

        \multicolumn{20}{@{}l@{}}{
            \textcolor{darkGreen}{\textit{Memory-based Methods}}
        }
        \\

        $\mu$VLA~\citep{cherepanov2026muvla}
        & 18.1
        & 60 & 0 & 4 & 0 & 0 & 12.8
        & 48 & 8 & 60 & 0 & 28 & 28.8
        & 0 & 0 & 40 & 20 & 4 & 12.8
        \\

        MME~\citep{dai2026robomme}
        & 18.4
        & 0 & 0 & 4 & 0 & 8 & 2.4
        & \textbf{100} & 4 & 56 & 0 & 16 & 35.2
        & 16 & 16 & 28 & 28 & 0 & 17.6
        \\

        SAM2Act+~\citep{fang2025sam2act}
        & \underline{51.2}
        & 32 & \underline{52} & \underline{32} & \textbf{48} & 68
        & \underline{46.4}
        & 8 & 28 & \underline{96} & \underline{28} & \underline{72}
        & 46.4
        & \underline{52} & 8 & \textbf{72} & \textbf{100}
        & \underline{72} & \underline{60.8}
        \\

        \textbf{SEEK}
        & \textbf{62.9}
        & \textbf{92} & 48 & \textbf{52} & 36 & \textbf{80}
        & \textbf{61.6}
        & 24 & \underline{32} & \textbf{100} & \textbf{40}
        & \textbf{100} & \textbf{59.2}
        & \textbf{60} & \textbf{32} & \underline{68} & \underline{88}
        & \textbf{92} & \textbf{68.0}
        \\

        \bottomrule
    \end{tabular}%
    }
\end{table*}

\subsection{Experimental Setup} %%%%%%%%%%%%%%%%%%%%%%%%%%%%%%%%%%%%%%%%%%%%%%%%%%%%%%%% sec 5.1 setip
\label{sec:exp_setup}

We primarily evaluate SEEK on HIDE, comprising 15 tasks across
Repetition Counting (RC), Historical-State Recall (HSR), and
Execution-Progress Tracking (EPT). We additionally evaluate standard
manipulation on 18-task RLBench, perturbation robustness on
\texttt{The Colosseum}, and physical execution on four real-robot tasks.
Baselines include vision-language-action models, 3D manipulation policies,
and memory-based methods, with SAM2Act as the backbone baseline.

Following RLBench's demonstration-generation and data-splitting protocol,
we use 100 training demonstrations and 25 held-out test episodes per
HIDE task. A single policy is jointly trained across all 15 tasks.
We report per-task success rates, category averages, and the overall
average, and ablate memory content, length, and components.
Simulation settings, baseline configurations, training procedures,
and real-robot protocols are detailed in \autoref{app:exp_setup}.

\subsection{Q1--Q2: Hidden-State Memory Gap and SEEK Performance}
\label{sec:q1_q2}

\autoref{tab:main_results} shows that HIDE remains challenging even
for memory-based policies. SAM2Act+ improves average success from
42.7\% for SAM2Act~\citep{fang2025sam2act} to 51.2\%, while
$\mu$VLA~\citep{cherepanov2026muvla} improves from 13.9\% for
OpenVLA-OFT~\citep{kim2025fine} to 18.1\%.
However, clear category-specific gaps remain: SAM2Act+ reaches 60.8\%
on EPT but only 46.4\% on RC and HSR, while
MME~\citep{dai2026robomme} achieves 35.2\% on HSR but only 2.4\% on RC.
The strongest baseline also varies by category, with SAM2Act+ leading
RC and EPT and GR00T-N1.7~\citep{bjorck2025gr00t} leading HSR at 49.6\%.

SEEK addresses these gaps more consistently, achieving 62.9\% average
success and outperforming the strongest baseline, SAM2Act+, by
11.7 percentage points. It leads all three categories with 61.6\% on RC,
59.2\% on HSR, and 68.0\% on EPT, corresponding to gains of 15.2,
9.6, and 7.2 points over the strongest category-wise baselines.
These results show that existing memory mechanisms do not uniformly
address different hidden-state requirements, while SEEK improves
performance across all three.

%%%%%%%%%%%%%%%%%%%%%%%%%%%%%%%%%%%%%%%%%%%%%%%%%%%%%%%%%%%%%%%%%%%%%% figure bar+radar
\begin{figure*}[t]
    % \vspace{-5pt}
    \centering
    \begin{subfigure}[t]{0.60\textwidth}
        \centering
        \includegraphics[width=\linewidth]{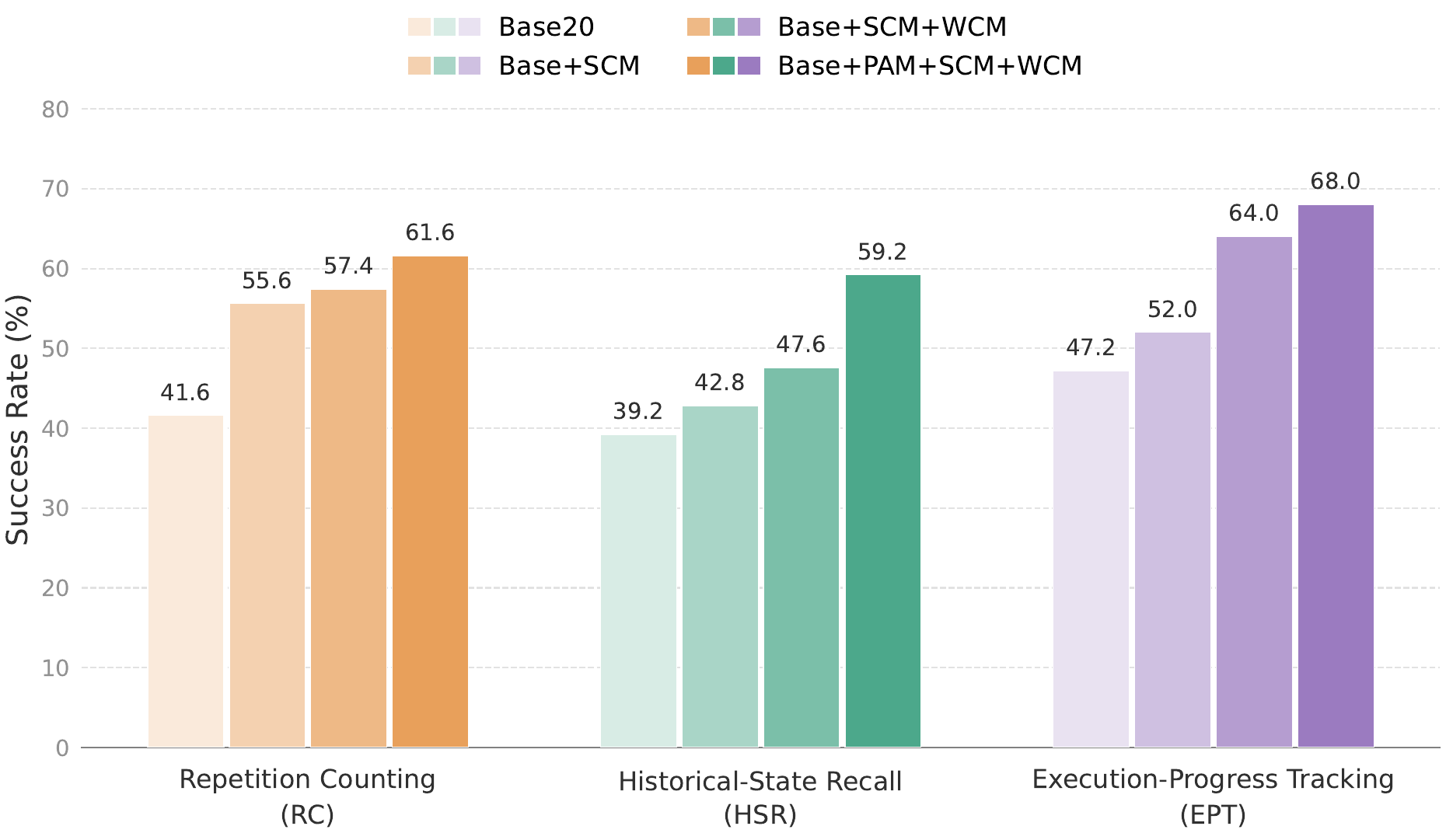}
        \caption{Progressive component integration.}
        \label{fig:module_bar}
    \end{subfigure}
    \hfill
    \begin{subfigure}[t]{0.38\textwidth}
        \centering
        \includegraphics[width=\linewidth]{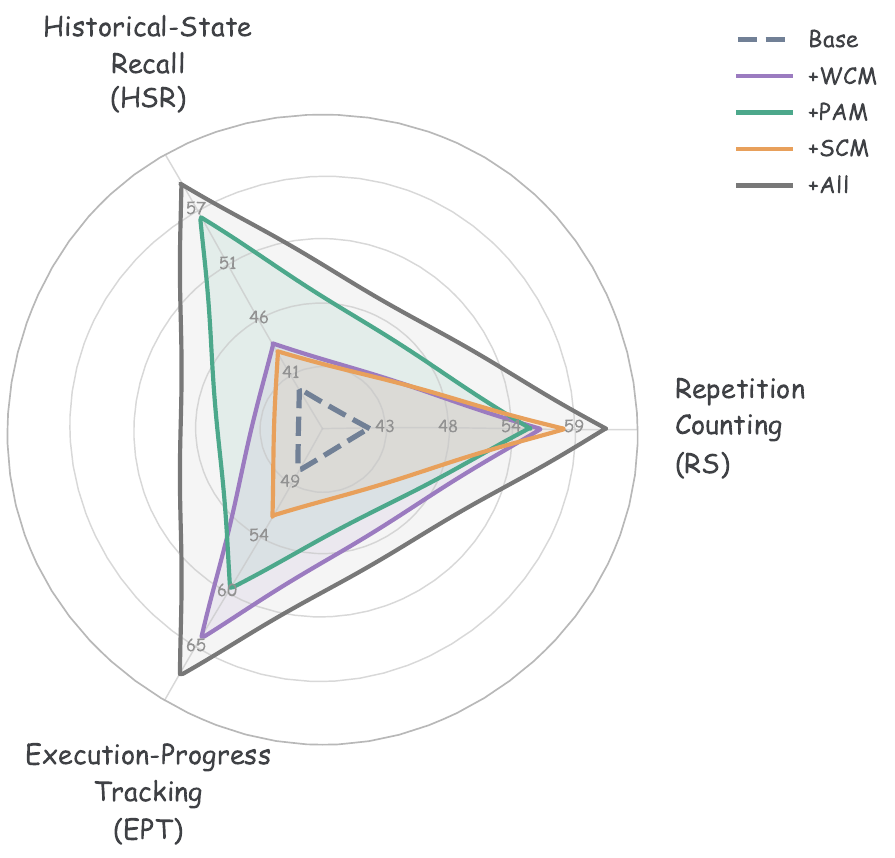}
        \caption{Individual and combined components.}
        \label{fig:module_radar}
    \end{subfigure}
    \caption{
        \textbf{Memory component analysis.}
        Category success rates (\%) for progressive integration (a)
        and individual components versus their combination (b).
        The components show complementary strengths across hidden-state
        requirements. Radar axes use category-specific scales.
    }
    \label{fig:module_analysis}
    % \vspace{-5pt}
\end{figure*}

\subsection{Q3: How Do Different Memory Components Contribute?} %%%%%%%%%%%%%%%%%%%%%%%%%%%%%%%%%% Q3 ablation
\label{sec:q3}

We examine what information to store, how much history to retain,
and how individual memory components contribute to performance.

\paragraph{What to store?}
With encoding dimensionality and memory architecture fixed, storing
visual features raises average success from 42.5\% to 62.9\%
(\autoref{tab:memory_design}). Gains span all three categories and
are largest on RC (+30.4 percentage points), followed by HSR (+17.6)
and EPT (+13.4). This suggests that visual histories retain
more task-relevant information than proprioceptive histories for
resolving hidden-state ambiguity in this setting.

%%%%%%%%%%%%%%%%%%%%%%%%%%%%%%%%%%%%%%%%%%%%%%%%%%%%%%%%%%%%%%%% tab:mem content+length ablation
% \begin{wraptable}{r}{0.48\linewidth}
%     \vspace{-12pt}
%     \centering
%     \caption{
%         \textbf{Memory content and length ablation.}
%         Success rates are in \%.
%         Content variants share the same encoding dimensionality and memory architecture.
%         % RC: Repetition Counting; HSR: Historical-State Recall;
%         % EPT: Execution-Progress Tracking.
%     }
%     \label{tab:memory_design}

%     \small
%     \setlength{\tabcolsep}{6pt}
%     \renewcommand{\arraystretch}{1.15}

%     \resizebox{\linewidth}{!}{%
%     \begin{tabular}{l c cccc}
%         \toprule
%         & \textbf{Variant}
%         & \textbf{Avg.}
%         & \textbf{RC}
%         & \textbf{HSR}
%         & \textbf{EPT} \\
%         \midrule

%         \multirow{2}{*}{\makecell[l]{Memory\\Content}}
%         & Proprioception & 42.5 & 31.2 & 41.6 & 54.6 \\
%         & Visual Features & \textbf{62.9} & \textbf{61.6} & \textbf{59.2} & \textbf{68.0} \\

%         \midrule

%         \multirow{4}{*}{\makecell[l]{Memory\\Length}}
%         & 0 & 46.7 & 46.4 & 42.0 & 51.6 \\
%         & 2 & 53.7 & 44.2 & 51.8 & 65.0 \\
%         & 4 & {57.3} & {48.6} & {55.8} & {68.2} \\
%         % &  6 & \textbf{60.3} & \textbf{54.2} & \textbf{57.8} & \textbf{69.0} \\
%         & 6 & \textbf{62.9} & \textbf{61.6} & \textbf{59.2} & \textbf{68.0} \\

%         \bottomrule
%     \end{tabular}
%     }%resize-end
%     \vspace{-16pt}
% \end{wraptable}
\providecommand{\samreported}[1]{\textcolor{gray}{#1}}

\begin{table*}[t]
    \centering

    % ==================== 左表：Memory ablation ====================
    \begin{minipage}[t]{0.48\linewidth}
        \centering
        \caption{
            \textbf{Memory content and length ablation.}
            Success rates (\%).
            Content variants share the same encoding dimensionality
            and memory architecture.
        }
        \label{tab:memory_design}
        \vspace{-5pt}

        \small
        \setlength{\tabcolsep}{5pt}
        \renewcommand{\arraystretch}{1.15}

        \resizebox{\linewidth}{!}{%
        \begin{tabular}{@{}lccccc@{}}
            \toprule
            & \textbf{Variant}
            & \textbf{Avg.}
            & \cellcolor{HIDEOrange!20}\textbf{RC}
            & \cellcolor{HIDEGreen!20}\textbf{HSR}
            & \cellcolor{HIDEPurple!20}\textbf{EPT} \\
            \midrule

            \multirow{2}{*}{\makecell[l]{Memory\\Content}}
            & Proprioception
            & 42.5 & 31.2 & 41.6 & 54.6 \\

            & Visual Features
            & \textbf{62.9}
            & \textbf{61.6}
            & \textbf{59.2}
            & \textbf{68.0} \\

            \midrule

            \multirow{4}{*}{\makecell[l]{Memory\\Length}}
            & 0
            & 46.7 & 46.4 & 42.0 & 51.6 \\

            & 2
            & 53.7 & 44.2 & 51.8 & 65.0 \\

            & 4
            & 57.2 & 48.6 & 55.8 & 67.2 \\

            & 6
            & \textbf{62.9}
            & \textbf{61.6}
            & \textbf{59.2}
            & \textbf{68.0} \\

            \bottomrule
        \end{tabular}%
        }
    \end{minipage}\hfill%
    % ==================== 右表：RLBench / Colosseum ====================
    \begin{minipage}[t]{0.48\linewidth}
    \vspace{0pt}
    \centering
    \caption{
        \textbf{RLBench and \texttt{The Colosseum} results.}
        Success rates (\%); parentheses give relative drops from Clean.
        $^{**}$ denotes our reproduction.
        Details: \autoref{app:detailed_results}.
    }
    \label{tab:rlbench_colosseum_summary}

    \small
    \setlength{\tabcolsep}{5pt}
    \renewcommand{\arraystretch}{1.15}

    \resizebox{\linewidth}{!}{%
    \begin{tabular}{@{}lccc@{}}
        \toprule
        \multirow{2}{*}{\textbf{Method}}
        & \textbf{RLBench}
        & \multicolumn{2}{c@{}}{\textbf{\texttt{The Colosseum}}} \\
        \cmidrule(lr){2-2}
        \cmidrule(l){3-4}

        & \textbf{Avg. Success} $\uparrow$
        & \textbf{Clean} $\uparrow$
        & \textbf{Average} $\uparrow$ \\
        \midrule

        RVT
        & 62.9
        & 43.6
        & 36.3 ($\downarrow$16.7\%) \\

        RVT-2
        & 81.4
        & 67.8
        & 59.5 ($\downarrow$12.3\%) \\

        \midrule

        SAM2Act$^{**}$
        & 84.1
        & \underline{68.4}
        & 61.5 ($\downarrow$10.1\%) \\

        \rowcolor{blue!6}
        SEEK
        & \textbf{84.7}
        & \textbf{68.6}
        & \textbf{61.9} ($\downarrow$9.8\%) \\

        \bottomrule
    \end{tabular}%
    }
\end{minipage}

\end{table*}

\paragraph{How much history to retain?}
Average success increases with memory length, rising from 46.7\%
at length 0 to 62.9\% at length 6 (\autoref{tab:memory_design}).
Length 6 achieves the best overall performance and the highest scores
on all three categories. Notably, RC initially drops from 46.4\% to
44.2\% at length 2, suggesting that a short memory horizon is still
insufficient for repetition-based tasks. With longer context, RC
improves substantially to 61.6\%. EPT shows a more gradual saturation,
increasing from 67.2\% at length 4 to 68.0\% at length 6.
Overall, these results indicate that sufficiently long temporal context
is important for reliably resolving hidden states, especially for
repetition counting.

\paragraph{Memory components.}
Progressively adding SCM, WCM, and PAM improves success across
all three categories (\autoref{fig:module_bar}). The largest gain
at each step occurs on RC with SCM (+14.0 percentage points),
EPT with WCM (+12.0), and HSR with PAM (+11.6).
The single-component comparison (\autoref{fig:module_radar})
also shows distinct strengths: SCM leads the individual variants
on RC, PAM on HSR, and WCM on EPT, while their combination leads
all three categories. This pattern is consistent with complementary
roles in all category of tasks.

\subsection{Q4: Does the Method Remain Effective Beyond HIDE?} %%%%%%%%%%%%%%%%%%%%%%%%%%%%% Q4
\label{sec:q5}

We further evaluate SEEK on standard manipulation tasks, under
environmental perturbations, and in real-world execution to assess
whether its memory-based design remains effective beyond HIDE.

\paragraph{Standard manipulation.}
SEEK achieves 84.7\% average success on RLBench
(\autoref{tab:rlbench_colosseum_summary}), slightly improving over our
SAM2Act~\citep{fang2025sam2act} reproduction at 84.1\%, and outperforming
RVT-2 and RVT at 81.4\% and 62.9\%, respectively.
This indicates that introducing memory preserves strong performance on
standard manipulation tasks rather than trading it off for hidden-state
reasoning.

\paragraph{Perturbation robustness.}
On \texttt{The Colosseum}, SEEK achieves 68.6\% on Clean and 61.9\%
averaged over perturbations, compared with 68.4\% and 61.5\% for our
SAM2Act reproduction (\autoref{tab:rlbench_colosseum_summary}).
The relative drop from Clean is also slightly smaller
(9.8\% vs.\ 10.1\%).
Together, these results show that SEEK maintains competitive robustness
under environmental perturbations while retaining its gains on
hidden-state tasks.

\paragraph{Real-world execution.}
%%%%%%%%%%%%%%%%%%%%%%%%%%%%%%%%%%%%%%%%%%%%%%%%%%%%%%%%%%%%%%%%%%%%%%%%%% tab:real-world

\begin{table}
    \centering
    \caption{
        \textbf{Real-world results.}
        Success rates (\%); Avg.\ is the unweighted mean across tasks.
    }
    \label{tab:real_world}
    \label{tab:in-out-dist} % Alias retained for existing cross-references.

    \small
    \setlength{\tabcolsep}{4.5pt}
    \renewcommand{\arraystretch}{1.15}

    % \resizebox{0.99\linewidth}{!}{
    \begin{tabular}{lccccc}
        \toprule
        \textbf{Method}
        & \textbf{Avg.}
        & \textbf{Push button $\times$ N}
        & \textbf{Stack N cups (M total)}
        & \textbf{Clean desk}
        & \textbf{Search chip} \\
        \midrule

        $\pi_{0.5}$%~\citep{physicalintelligence2025pi05}
        & 13 & 0 & 20 & 0 & 32 \\

        SAM2Act+%~\citep{fang2025sam2act}
        & 47 & 32 & 48 & 44 & 64 \\

        \rowcolor{blue!6}
        SEEK
        & \textbf{89}
        & \textbf{76}
        & \textbf{100}
        & \textbf{80}
        & \textbf{100} \\

        \bottomrule
    \end{tabular}
\end{table}

We evaluate four real-world tasks: repeated button pressing, cup stacking,
desk cleaning, and searching for a hidden white piece, each under randomized
initial configurations. Task details are provided in
Appendix~\ref{app:real-tasks}. These tasks contain visually similar
observations associated with different latent states and actions, requiring
task history for correct execution. SEEK achieves 89\% average success
(\autoref{tab:real_world}), compared with 47\% for
SAM2Act+~\citep{fang2025sam2act} and 13\% for
$\pi_{0.5}$~\citep{physicalintelligence2025pi05}, supporting the use of
memory to track initial state, task progress, and recent scene context.
\section{Conclusion}

We introduced HIDE, a benchmark for evaluating hidden-state reasoning in robotic manipulation, and SEEK, a memory framework that maintains task-relevant information over interaction history. HIDE decomposes hidden-state manipulation into three complementary requirements---repetition counting, historical-state recall, and execution-progress tracking---and exposes clear limitations in existing policies, including methods equipped with memory. SEEK consistently improves performance across these categories, while retaining strong performance on standard manipulation tasks, under environmental perturbations, and in real-world execution. Our ablations further show that effective memory depends not only on retaining history, but also on what information is stored, how long it is preserved, and how different memory mechanisms contribute to distinct hidden-state requirements. Together, these results highlight the importance of explicitly modeling latent task state for long-horizon robotic decision making. HIDE currently focuses on controlled and interpretable forms of hidden state; extending it to richer real-world settings, broader sources of partial observability, and more general memory mechanisms remains an important direction for future work.

\newpage
\subsection*{AI use statement}
In this work, we used generative AI tools for (i) literature search and
summarization of related work, (ii) scripting and queueing the batch
execution of experiments, and (iii) drafting the initial skeleton of parts
of the manuscript and figure layouts. We have not used generative AI tools
for research ideation, experimental design, code implementation and
debugging, data construction and analysis, or the final writing and revision of the
manuscript, and AI-assisted data collection and AI-assisted peer review are
not applicable to this work. Additionally, we used generative AI tools for
language polishing and LaTeX formatting. We have reviewed all AI-assisted
work: all references suggested by AI-assisted literature summarization were
manually verified against the original papers; experiment execution scripts
were inspected by the authors and all reported results were independently
checked; all AI-drafted text and figures were reviewed, rewritten, and
finalized by the authors. We take responsibility for the final content of
this work, including text, claims or artifacts produced with the aid of
generative AI.

\subsection*{Ethics statement}
This work does not involve human subjects and required no IRB approval. All
simulation experiments are conducted on publicly available, open-source
benchmarks (RLBench~\citep{james2020rlbench} and The Colosseum~\citep{pumacay2024colosseum}), used in accordance with their
licenses; these benchmarks consist of synthetic scenes and contain no
personally identifiable information. Real-robot evaluations are conducted in
controlled indoor laboratory environments under human supervision and
standard safety protocols, and no personal data of bystanders is collected
or stored. We do not foresee direct societal harms, dual-use risks, or
fairness concerns beyond those common to robotic manipulation research in
general.

% \subsection*{Reproducibility statement}

% (This section is \textbf{recommended} and does not count toward the page limit.)

% It is important that the work published in ICLR is reproducible. Authors are
% strongly encouraged to include a paragraph-long Reproducibility Statement at the
% end of the main text (before references) to discuss the efforts that have been
% made to ensure reproducibility. This paragraph should not itself describe
% details needed for reproducing the results, but rather reference the parts of
% the main paper, appendix, and supplemental materials that will help with
% reproducibility. For example, for novel models or algorithms, a link to an
% anonymous downloadable source code can be submitted as supplementary materials;
% for theoretical results, clear explanations of any assumptions and a complete
% proof of the claims can be included in the appendix; for any datasets used in
% the experiments, a complete description of the data processing steps can be
% provided in the supplementary materials. Each of the above are examples of
% things that can be referenced in the reproducibility statement.


\section*{Acknowledgements}
This work was completed at the Shanghai Artificial Intelligence Laboratory. We gratefully acknowledge the computational resources provided by the Shanghai Artificial Intelligence Lab, which made this research possible.

% \newpage
\bibliography{main}
\bibliographystyle{iclr2027_conference}

\newpage
\appendix
\section*{Appendix}

% Insert within your existing appendix.
\section{Task Details and Demonstration}
\label{app:hide_task_descriptions}

\subsection{simulation tasks}
\label{app:sim_tasks}
HIDE contains 15 tasks organized into three categories with five tasks
in each: repetition counting, historical-state recall, and execution-progress tracking. 
Brief statistics are displayed in~\cref{tab:task_stats}.

%%%%%%%%%%%%%%%%%%%%%%%%%%%%%%%%%%%%%%%%%%%%%%%%%%%%%%%%%%%%%%%%%%%%%%%%%%%%%% 15 task stat
\begin{table*}[h!]
\centering
\caption{
\textbf{Task Statistics of HIDE.}
We report the language template, average number of frames,
average number of extracted keyframes,
number of task variations, and variation type.
Square brackets denote variable placeholders in the language templates,
with \texttt{[N]} indicating the requested count.
}
\label{tab:task_stats}
\scriptsize
\setlength{\tabcolsep}{2pt}
\renewcommand{\arraystretch}{1.15}

\begin{tabularx}{\textwidth}{@{}lQcccl@{}}
\toprule
Task name
& Language Template
& Frames
& Keyframes
& \# of Var.
& Variation Type \\
\midrule

push\_button\_times
& ``push the \texttt{[color]} button \texttt{[N]} times''
& 144.3 & 7.8 & 50
& repetition count $\times$ button color \\

stack\_blocks
& ``stack \texttt{[N]} \texttt{[color]} blocks''
& 372.8 & 16.1 & 60
& stack size $\times$ color \\

change\_channel\_times
& ``turn the channel \texttt{[direction]} \texttt{[N]} times''
& 244.5 & 12.2 & 6
& direction $\times$ repetition count \\

stack\_cups\_new
& ``stack \texttt{[N]} cup(s) on top of the \texttt{[color]} cup''
& 208.8 & 8.4 & 40
& base color $\times$ cup count \\

stack\_blocks\_new
& ``put \texttt{[N]} \texttt{[color]} blocks into the rectangular container''
& 283.5 & 11.8 & 60
& placement count $\times$ color \\

weighing\_on\_off
& ``weigh the pepper and put it in another container''
& 199.0 & 11.0 & 2
& starting container \\

pour\_put\_back
& ``pour liquid from the \texttt{[color]} cup to the \texttt{[color]} cup, then place it on the other coaster''
& 330.0 & 7.5 & 40
& coaster side $\times$ cup-color pair \\

wipe\_desk\_rubbish
& ``wipe dirt off the desk''
& 272.9 & 9.0 & 1
& randomized layout \\

light\_bulb\_in\_out
& ``screw in the light bulb and move it to the other holder''
& 335.9 & 10.8 & 20
& holder color (implicit) \\

reopen\_drawer
& ``close the opened drawer, push the button, and open the previous drawer again''
& 320.2 & 10.0 & 3
& drawer level (bottom/middle/top) \\

search\_drawer
& ``take item out of the drawer''
& 393.8 & 15.7 & 3
& hidden-item drawer level \\

search\_boxes
& ``take shoes out of box''
& 629.1 & 19.0 & 2
& search path / holder config \\

search\_cup\_from\_cabinet
& ``take out a cup from the cabinet''
& 245.5 & 9.9 & 2
& cabinet side \\

lift\_and\_check
& ``lift blocks one by one to find the white chip underneath and remove it''
& 277.3 & 14.3 & 80
& block color $\times$ chip position \\

swap\_square\_pegs
& ``swap the positions of the two rings''
& 321.3 & 18.0 & 6
& peg/ring permutation \\

\bottomrule
\end{tabularx}%
\end{table*}
%%%%%%%%%%%%%%%%%%%%%%%%%%%%%%%%%%%%%%%%%%%%%%%%%%%%%%%%%%%%%%%%%%%%%%%%%%%%%% end

\HIDEcategory
  {HIDEOrange}
  {I. Repetition Counting}
  {Track completed repetitions to know when to stop.}

% ------------------------------------------------------------
% (a)
% ------------------------------------------------------------
\begin{HIDEtask}{HIDEOrange}{(a) Push button times}

\HIDEfield{Task Description}{
The robot is instructed to identify the target colored button and
press it the requested number of times. Depending on the variation,
the button must be pressed once, twice, three times, four times,
or five times.
}

\HIDEfield{Success Metric}{
The task is considered successful once the target button has been
pressed exactly the requested number of times and the robot arm
has withdrawn upward.
}

\HIDEfield{Objects}{
Three colored push buttons and their button mechanisms.
}

\HIDEstats{50}{4, 6, 8, 10, or 12}

\HIDEinstruction{Push the \HIDEslot{color} button \HIDEslot{count}.}

\end{HIDEtask}

% ------------------------------------------------------------
% (b)
% ------------------------------------------------------------
\begin{HIDEtask}{HIDEOrange}{(b) Stack blocks}

\HIDEfield{Task Description}{
The robot is instructed to identify blocks of the specified color
and stack the requested number of them vertically. The requested
stack contains two, three, or four blocks.
}

\HIDEfield{Success Metric}{
The task is considered successful once the requested number of
target blocks is detected in the stacking region and the robot
is no longer holding an object.
}

\HIDEfield{Objects}{
Four target blocks, four distractor blocks, and a stacking region.
}

\HIDEstats{60}{11, 17, 22, 23, or 24}

\HIDEinstruction{Stack \HIDEslot{2\textendash4} \HIDEslot{color} blocks.}

\end{HIDEtask}

% ------------------------------------------------------------
% (c)
% ------------------------------------------------------------
\begin{HIDEtask}{HIDEOrange}{(c) Change channel times}

\HIDEfield{Task Description}{
The robot is instructed to pick up the television remote, point it
toward the television, and press either the plus or minus channel
button the requested number of times.
}

\HIDEfield{Success Metric}{
The task is considered successful once the correct channel button
has been pressed exactly one, two, or three times, the remote
reaches the required final configuration, and the robot arm has
withdrawn upward.
}

\HIDEfield{Objects}{
A television remote, plus and minus buttons, and a
television-facing target region.
}

\HIDEstats{6}{10, 12, or 14}

\HIDEinstruction{Turn the channel \HIDEslot{up or minus}
\HIDEslot{once, twice, or three times}.}

\end{HIDEtask}

% ------------------------------------------------------------
% (d)
% ------------------------------------------------------------
\begin{HIDEtask}{HIDEOrange}{(d) Stack cups new}

\HIDEfield{Task Description}{
The robot is instructed to keep the specified colored cup as
the base and place either one or two of the remaining cups
on top of it.
}

\HIDEfield{Success Metric}{
The task is considered successful once the requested cup or cups
are detected in the stack, the gripper is empty, and the robot
arm has withdrawn upward.
}

\HIDEfield{Objects}{
Three colored cups and a cup-stacking success region.
}

\HIDEstats{40}{6 or 11}

\HIDEinstruction{Stack \HIDEslot{one or two} cup(s) on top of
the \HIDEslot{color} cup.}

\end{HIDEtask}

% ------------------------------------------------------------
% (e)
% ------------------------------------------------------------
\begin{HIDEtask}{HIDEOrange}{(e) Stack blocks new}

\HIDEfield{Task Description}{
The robot is instructed to identify blocks of the specified color
and place the requested number of them into the open rectangular
container at the center of the workspace.
}

\HIDEfield{Success Metric}{
The task is considered successful once the requested number of
target blocks is detected inside the container, the gripper is
empty, and the robot arm has withdrawn upward.
}

\HIDEfield{Objects}{
Four target blocks, four distractor blocks, and a central
open-top container.
}

\HIDEstats{60}{6, 11, 16, or 17}

\HIDEinstruction{Put \HIDEslot{1\textendash3} \HIDEslot{color}
blocks into the rectangular container in the center.}

\end{HIDEtask}

\HIDEcategory
  {HIDEGreen}
  {II. Historical-State Recall}
  {Act on past information no longer visible in the scene.}

% ------------------------------------------------------------
% (f)
% ------------------------------------------------------------
\begin{HIDEtask}{HIDEGreen}{(f) Weighing on off}

\HIDEfield{Task Description}{
The robot is instructed to pick up the pepper from one container,
place it on the scale long enough to obtain a weight reading,
and then move it into the other container.
}

\HIDEfield{Success Metric}{
The task is considered successful once the scale has registered
the pepper while the gripper is released and the pepper is
subsequently detected in the destination container.
}

\HIDEfield{Objects}{
A pepper, a weighing scale, and two containers.
}

\HIDEstats{2}{11}

\HIDEinstruction{Weigh the pepper and put it in another container.}

\end{HIDEtask}

% ------------------------------------------------------------
% (g)
% ------------------------------------------------------------
\begin{HIDEtask}{HIDEGreen}{(g) Pour put back}

\HIDEfield{Task Description}{
The robot is instructed to pick up the source cup, pour its
liquid into the target cup, and then place the source cup on
the other coaster.
}

\HIDEfield{Success Metric}{
The task is considered successful once all liquid particles are
detected in the target cup and the source cup is detected on
the opposite coaster.
}

\HIDEfield{Objects}{
A source cup, a target cup, liquid particles, and two coasters.
}

\HIDEstats{40}{6, 7, or 8}

\HIDEinstruction{Pour liquid from the \HIDEslot{source color} cup
to the \HIDEslot{target color} cup, then place the source cup
on the other coaster.}

\end{HIDEtask}

% ------------------------------------------------------------
% (h)
% ------------------------------------------------------------
\begin{HIDEtask}{HIDEGreen}{(h) Wipe desk rubbish}

\HIDEfield{Task Description}{
The robot is instructed to dispose of the rubbish in the bin
and then use the sponge to wipe all visible dirt from the desk.
}

\HIDEfield{Success Metric}{
The task is considered successful once the rubbish is detected
in the bin and all generated dirt spots have been removed
from the desk.
}

\HIDEfield{Objects}{
A piece of rubbish, a rubbish bin, a sponge, and multiple
dirt spots.
}

\HIDEstats{1}{8 or 9}

\HIDEinstruction{Wipe dirt off the desk.}

\end{HIDEtask}

% ------------------------------------------------------------
% (i)
% ------------------------------------------------------------
\begin{HIDEtask}{HIDEGreen}{(i) Light bulb in out}

\HIDEfield{Task Description}{
The robot is instructed to pick up the light bulb from its
initial holder, screw it into the lamp until it lights,
remove it, and place it into the other holder.
}

\HIDEfield{Success Metric}{
The task is considered successful once the bulb has been detected
in the lamp, has lit up, is subsequently detected in the
destination holder, and has been released by the gripper.
}

\HIDEfield{Objects}{
A light bulb, a lamp socket, and two bulb holders.
}

\HIDEstats{20}{10 or 11}

\HIDEinstruction{Screw in the light bulb and move it to
the other holder.}

\end{HIDEtask}

% ------------------------------------------------------------
% (j)
% ------------------------------------------------------------
\begin{HIDEtask}{HIDEGreen}{(j) Reopen drawer}

\HIDEfield{Task Description}{
The robot is instructed to remember the drawer slot that was
initially open, close it, press the button on the table, and
then find and reopen the previously opened drawer.
}

\HIDEfield{Success Metric}{
The task is considered successful once the initially opened
drawer has been closed, the button has been pressed, and
the same drawer has been reopened.
}

\HIDEfield{Objects}{
A three-level drawer cabinet and a push button.
}

\HIDEstats{3}{10}

\HIDEinstruction{Close the drawer, then reopen the previously
opened drawer while pushing the button in between.}

\end{HIDEtask}

\HIDEcategory
  {HIDEPurple}
  {III. Execution-Progress Tracking}
  {Use history to identify which substeps are done and what comes next.}

% ------------------------------------------------------------
% (k)
% ------------------------------------------------------------
\begin{HIDEtask}{HIDEPurple}{(k) Search drawer}

\HIDEfield{Task Description}{
The robot is instructed to search the drawers, locate the
hidden item, remove it from the correct drawer, and place
it in the target region.
}

\HIDEfield{Success Metric}{
The task is considered successful once the hidden item is
detected in the designated success region outside the drawer.
}

\HIDEfield{Objects}{
A three-level drawer cabinet, a hidden item, and a success region.
}

\HIDEstats{3}{9, 16, or 23}

\HIDEinstruction{Take the item out of the drawer.}

\end{HIDEtask}

% ------------------------------------------------------------
% (l)
% ------------------------------------------------------------
\begin{HIDEtask}{HIDEPurple}{(l) Search boxes}

\HIDEfield{Task Description}{
The robot is instructed to open the shoe box, search for
both shoes, remove them, and place them on the table.
}

\HIDEfield{Success Metric}{
The task is considered successful once both shoes are detected
outside the box in the target region and the robot is no longer
holding either shoe.
}

\HIDEfield{Objects}{
A shoe box and lid, two shoes, and a target region on the table.
}

\HIDEstats{2}{15, 16, 23, or 24}

\HIDEinstruction{Take the shoes out of the box.}

\end{HIDEtask}

% ------------------------------------------------------------
% (m)
% ------------------------------------------------------------
\begin{HIDEtask}{HIDEPurple}{(m) Search cup from cabinet}

\HIDEfield{Task Description}{
The robot is instructed to open the appropriate side of the
cabinet, locate the cup, remove it from the cabinet, and
release it outside.
}

\HIDEfield{Success Metric}{
The task is considered successful once the cup is no longer
detected inside the cabinet and the robot is no longer
holding it.
}

\HIDEfield{Objects}{
A two-sided cabinet, sliding cabinet doors, and a cup.
}

\HIDEstats{2}{7, 8, 13, or 14}

\HIDEinstruction{Take out a cup from the cabinet.}

\end{HIDEtask}

% ------------------------------------------------------------
% (n)
% ------------------------------------------------------------
\begin{HIDEtask}{HIDEPurple}{(n) Lift and check}

\HIDEfield{Task Description}{
The robot is instructed to lift the blocks one by one,
inspect the space underneath each block, locate the hidden
white chip, and move the block that covers it away.
}

\HIDEfield{Success Metric}{
The task is considered successful once the block covering
the hidden white chip has been identified and moved into
the designated success region.
}

\HIDEfield{Objects}{
Four colored blocks, a hidden white chip or marker,
and a success region.
}

\HIDEstats{80}{6, 11, 16, or 21}

\HIDEinstruction{Lift blocks one by one to find the white chip
underneath and remove it.}

\end{HIDEtask}

% ------------------------------------------------------------
% (o)
% ------------------------------------------------------------
\begin{HIDEtask}{HIDEPurple}{(o) Swap square pegs}

\HIDEfield{Task Description}{
The robot is instructed to pick up the two square rings
and exchange their positions between the pegs.
}

\HIDEfield{Success Metric}{
The task is considered successful once each square ring is
detected on the other ring's original target peg.
}

\HIDEfield{Objects}{
Two square rings and three pegs.
}

\HIDEstats{6}{18}

\HIDEinstruction{Swap the positions of the two rings.}

\end{HIDEtask}

%%%%%%%%%%%%%%%%%%%%%%%%%%%%%%%%%%%%%%%%%%%%%%%%%%%%%%%%%%%%%%%%%%%%%%%%%%%%%%%%%%%%% secA.2 real-world

\subsection{Real-world Tasks}%%%%%%%%%%%%%%%%%%%%%%%%%%%%%%%%%%%%%%%%%%%%%%%%%%%%%%%%%%%%%%%%%
\label{app:real-tasks}

We design four real-world tasks that require reasoning over hidden task
states: repeated button pressing, cup stacking, desk cleaning, and
searching for a hidden white piece. In each task, visually similar
observations can correspond to different latent states and therefore
require different actions, making the tasks challenging for policies
that rely primarily on the current observation.

%%%%%%%%%%%%%%%%%%%%%%%%%%%%%%%%%%%%%%%%%%%%%%%%%%%%%%%%%%%%%%%%%%%%%%%%%% tab-Real-world task stat
\begin{table*}[h]
    \centering
    \caption{
        \textbf{Real-world task statistics.}
        Language templates, trajectory statistics, and variation types
        for the four real-world tasks.
    }
    \label{tab:hide_four_task_statistics}
    \resizebox{\linewidth}{!}{
    \begin{tabular}{@{}llcccl@{}}
        \toprule
        \textbf{Task}
        & \textbf{Language Template}
        & \textbf{Frames}
        & \textbf{Keyframes}
        & \textbf{\# Var.}
        & \textbf{Variation Type} \\
        \midrule

        push\_button\_times
        & ``push the \texttt{[color]} button \texttt{[N]} times''
        & 144.3 & 7.8 & 3
        & repetition count $\times$ randomized object placement \\

        stack\_cups
        & ``stack \texttt{[N]} cups on the middle cup''
        & 372.8 & 16.1 & 4
        & target stack count $\times$ randomized object placement \\

        wipe\_desk\_rubbish
        & ``clean the desk''
        & 272.9 & 9.0 & 2
        & rubbish location $\times$ randomized object placement \\

        lift\_and\_check
        & ``lift the blocks and look for the white piece''
        & 277.3 & 14.3 & 3
        & hidden-piece location $\times$ randomized object placement \\

        \bottomrule
    \end{tabular}
    }
\end{table*}

%%%%%%%%%%%%%%%%%%%%%%%%%%%%%%%%%%%%%%%%%%%%%%%%%%%%%%%%%%%%%%%%%%%%%%%%%%%%%% end

\HIDEcategory
  {HIDEGray}
  {I. Real-world Tasks}
  {Physical manipulation tasks requiring hidden-state reasoning.}

% ------------------------------------------------------------
% (a)
% ------------------------------------------------------------
\begin{HIDEtask}{HIDEGray}{(a) Push button multiple times}

\HIDEfield{Task Description}{
The robot is instructed to identify the target colored button and
press it a specified number of times. Since observations before and
after individual presses can be visually similar, the robot must track
the number of completed presses and determine whether to continue or stop.
}

\HIDEfield{Success Metric}{
The task is successful if the target button is pressed exactly the
requested number of times and the robot terminates without an additional
press.
}

\HIDEfield{Objects}{
Three colored push buttons and their corresponding button mechanisms.
}

\HIDEstats{3}{4, 6 or 8}

\HIDEinstruction{Push the \HIDEslot{color} button
\HIDEslot{count} times.}

\end{HIDEtask}

% ------------------------------------------------------------
% (b)
% ------------------------------------------------------------
\begin{HIDEtask}{HIDEGray}{(b) Stack cups}

\HIDEfield{Task Description}{
The robot is instructed to stack a specified number of cups on the
middle cup. Similar intermediate configurations may correspond to
different stages of task completion, requiring the robot to track
how many cups have already been stacked.
}

\HIDEfield{Success Metric}{
The task is successful once the requested number of cups has been
correctly stacked on the middle cup and the robot is no longer
holding a cup.
}

\HIDEfield{Objects}{
Paper cups arranged around a designated middle cup.
}

\HIDEstats{4}{11, 17 or 23}

\HIDEinstruction{Stack \HIDEslot{count} cups on the middle cup.}

\end{HIDEtask}

% ------------------------------------------------------------
% (c)
% ------------------------------------------------------------
\begin{HIDEtask}{HIDEGray}{(c) Clean the desk}

\HIDEfield{Task Description}{
The robot must first remove the rubbish from the desk, which reveals
the area that needs to be cleaned. It then uses the cleaning tool to
remove the remaining dirt. The robot must remember earlier scene
information and track which substeps have already been completed.
}

\HIDEfield{Success Metric}{
The task is successful once the rubbish has been removed and all
required regions of the desk have been cleaned.
}

\HIDEfield{Objects}{
A piece of rubbish, a bin, a cleaning tool, dirt regions, and a desk.
}

\HIDEstats{2}{9}

\HIDEinstruction{Clean the desk.}

\end{HIDEtask}

% ------------------------------------------------------------
% (d)
% ------------------------------------------------------------
\begin{HIDEtask}{HIDEGray}{(d) Lift and check}

\HIDEfield{Task Description}{
The robot is instructed to inspect a set of blocks to locate a white
piece hidden underneath one of them. It lifts the blocks sequentially
until the target is found. Because previously inspected locations may
again appear visually similar, the robot must remember which blocks
have already been checked.
}

\HIDEfield{Success Metric}{
The task is successful once the hidden white piece has been located
and the required final action has been completed.
}

\HIDEfield{Objects}{
Multiple blocks and a white piece hidden underneath one of them.
}

\HIDEstats{3}{6, 11, 16, or 21}

\HIDEinstruction{
Lift the blocks one by one and find the white piece underneath.
}

\end{HIDEtask}
%%%%%%%%%%%%%%%%%%%%%%%%%%%%%%%%%%%%%%%%%%%%%%%%%%%%%%%%%%%%%%%%%%%%%%%%%%%%
\subsection{Task Visualization}
\label{app:task_visual}

\begin{figure*}[h]
    \centering
    \includegraphics[width=0.99\linewidth]{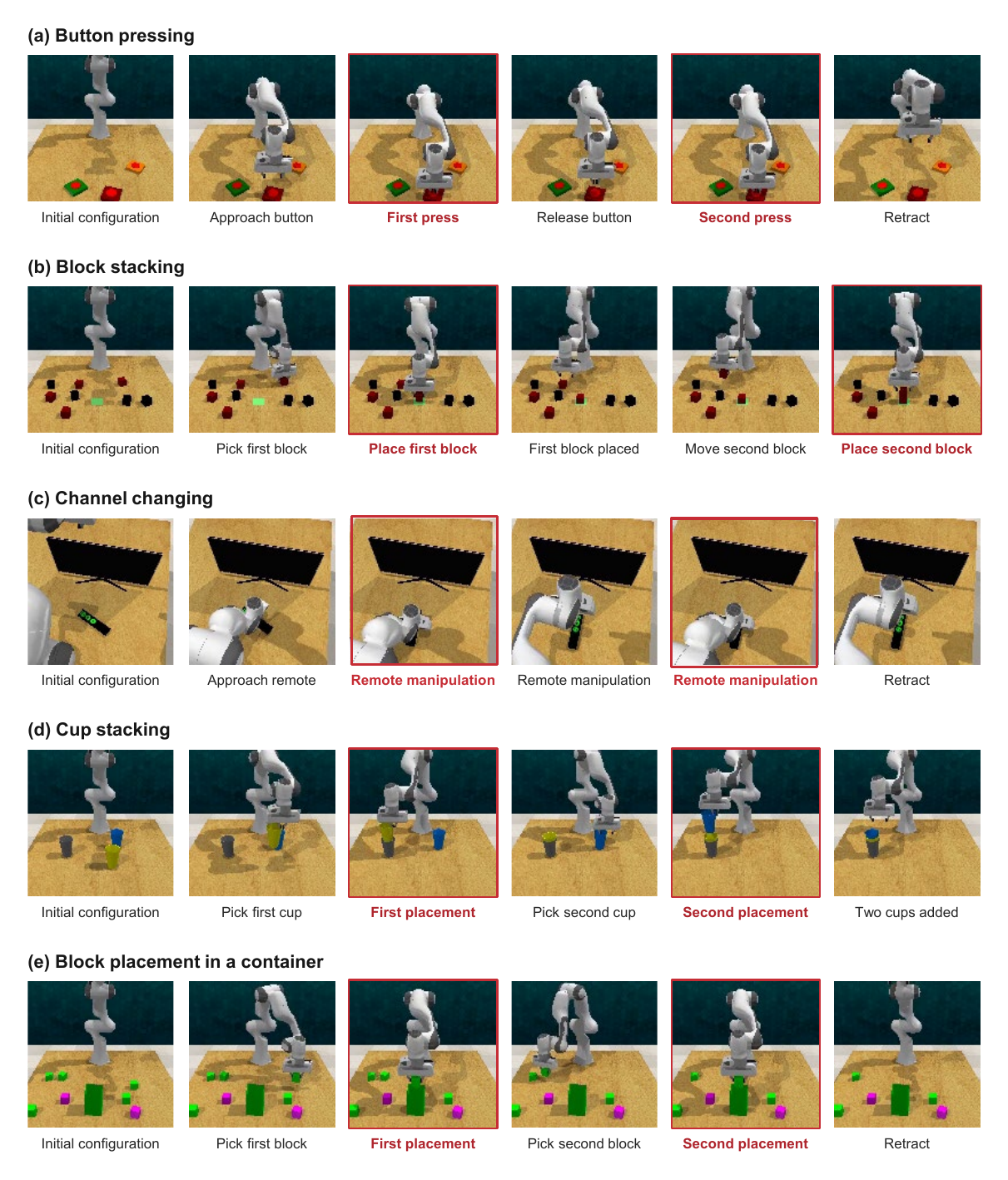}
    \caption{
        Illustration of Repetition Counting (RS) tasks.
    }
    \label{fig:traj1}
\end{figure*}

\begin{figure*}[p]
    \centering
    \includegraphics[width=0.99\linewidth]{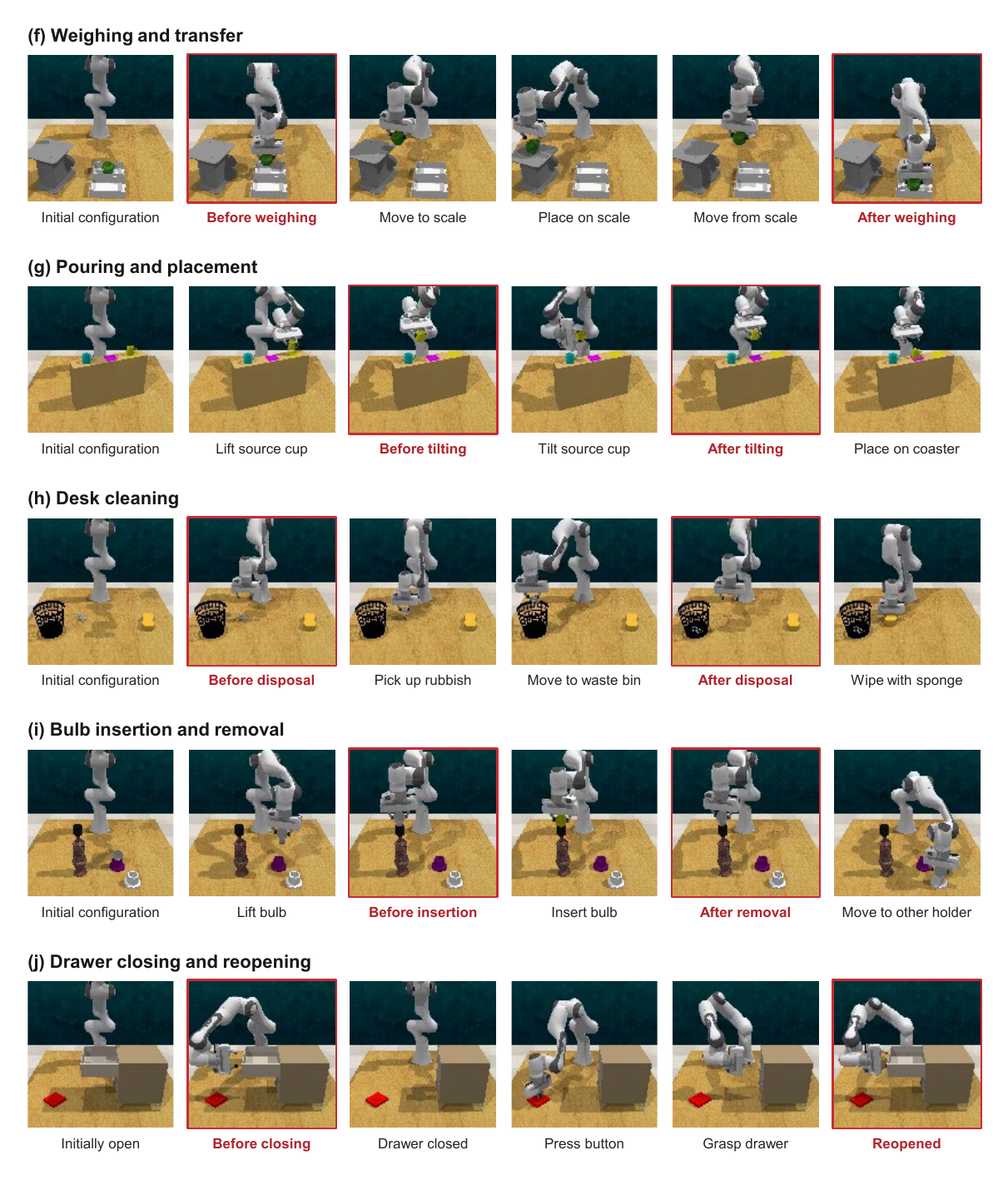}
    \caption{
        Illustration of Historical-State Recall (HSR) tasks.
    }
    \label{fig:traj2}
\end{figure*}

\begin{figure*}[p]
    \centering
    \includegraphics[width=0.99\linewidth]{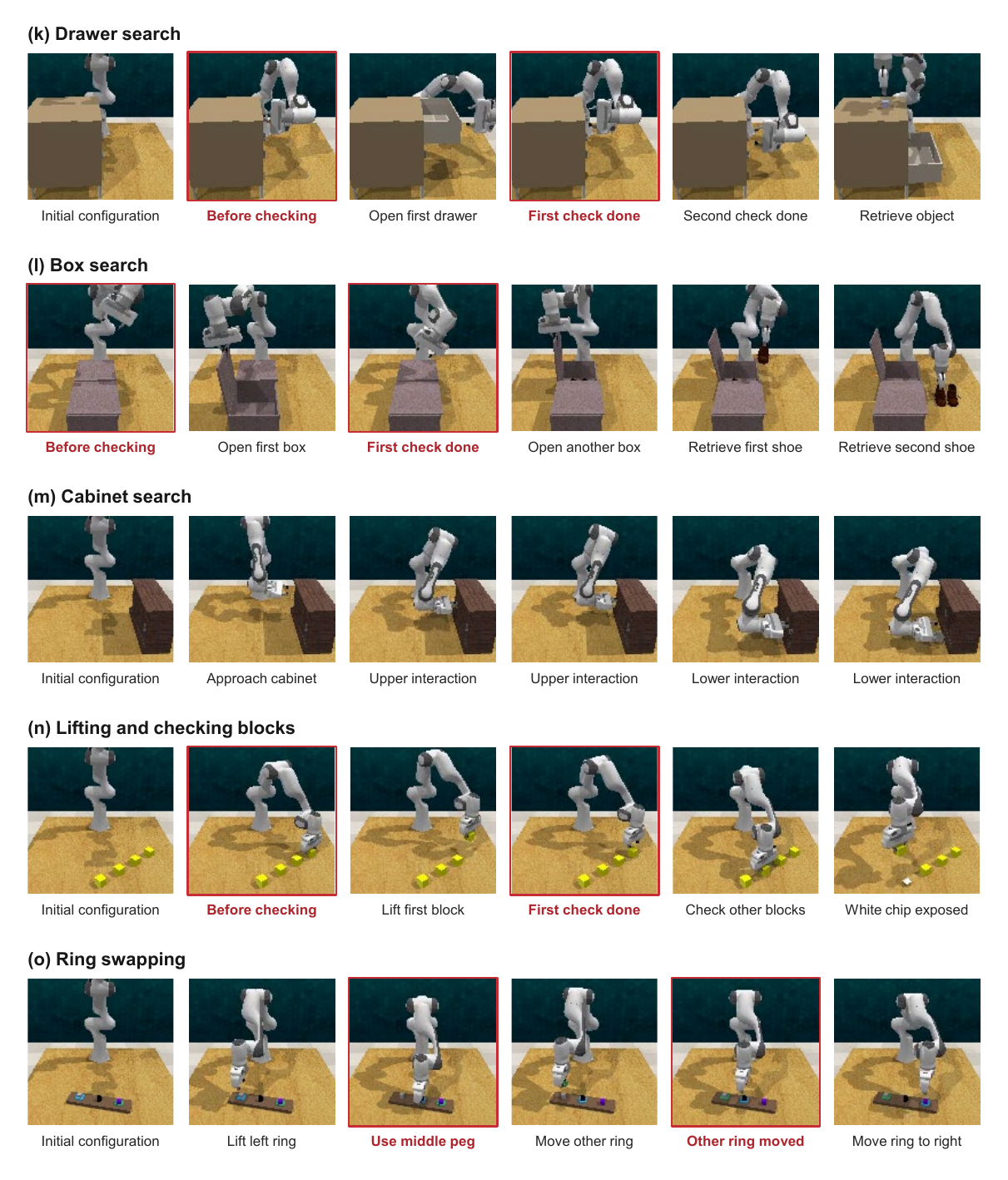}
    \caption{
        Illustration of Execution-Progress Tracking (EPT) tasks.
    }
    \label{fig:traj3}
\end{figure*}

Here we provide the complete visual sequences of all HIDE tasks in~\cref{fig:traj1,fig:traj2,fig:traj3}.
Red boxes highlight pairs of observations that are visually
similar while corresponding to different hidden task states. Although
the current visual appearance provides little information to
distinguish these states, the latent state determines the appropriate
next action. For example, visually similar scenes may correspond to
different repetition counts, different historically observed object
properties, or different execution stages, and therefore lead to
different action choices. These examples illustrate the central
challenge of HIDE: the current observation alone is insufficient to
determine the next action without recovering task-relevant information
from interaction history.

%%%%%%%%%%%%%%%%%%%%%%%%%%%%%%%%%%%%%%%%%%%%%%%%%%%%%%%%%%%%%%%%%%%%%%%%%%%%%%%%%%%%
\newpage
\section{Additional Experimental Details}
\label{app:exp_setup}

\subsection{Benchmarks and Data}
HIDE extends RLBench with 15 tasks organized into three categories---repetition counting, historical-state recall, and execution-progress tracking---with five tasks per category. We follow the RLBench workflow for demonstration generation, language-conditioned task specification, and train--test separation. For HIDE and the standard 18-task RLBench suite, we use 100 training demonstrations per task and 25 held-out test episodes per task. Training and test sets contain separate demonstration episodes with scene and task variations. HIDE serves as our principal benchmark for comparisons and component ablations, while the standard RLBench suite evaluates general manipulation performance. We use \texttt{The Colosseum} to assess robustness, reporting performance in clean scenes and under its benchmark-defined perturbations.

\subsection{Simulation and Observations}
The simulation experiments use CoppeliaSim through PyRep, with a 7-DoF Franka Emika Panda robot operating in a tabletop workspace. Our 3D policy uses RGB-D observations from four cameras: front, left shoulder, right shoulder, and wrist, each at $128\times128$ resolution. Calibrated observations are combined into a point cloud, together with the language instruction and robot proprioception. The policy predicts keyframe actions comprising a target end-effector position, orientation, and gripper state. An OMPL-based motion planner executes the corresponding motion to the target pose.

Our HIDE comparison covers three families of policies.
Vision-language-action models include
OpenVLA~\citep{kim2024openvla},
OpenVLA-OFT~\citep{kim2025fine},
$\pi_0$~\citep{black2024pi0},
$\pi_{0.5}$~\citep{physicalintelligence2025pi05}, and
GR00T-N1.7~\citep{bjorck2025gr00t}.
We also compare against 3D manipulation policies,
including RVT~\citep{goyal2023rvt},
RVT-2~\citep{goyal2024rvt}, and
SAM2Act~\citep{fang2025sam2act}.
Memory-based baselines include
$\mu$VLA~\citep{cherepanov2026muvla},
MME~\citep{dai2026robomme}, and SAM2Act+.
SAM2Act serves as the backbone reference for studying the effect of
introducing memory.
For external benchmarks, results from official weights, our reproductions,
and published reports are identified separately.

\paragraph{Training protocol.}
We jointly train a SEEK on all 15 HIDE tasks using the two-stage
procedure summarized in \autoref{appendix_tab:hyperparam_seek_hide}.
Stage~1 learns a memory-free manipulation policy, adapting the pretrained
visual backbone through LoRA while optimizing the coarse and fine action
branches. Stage~2 starts from the Stage~1 epoch-20 checkpoint and introduces
temporal memory training. The visual backbone, including its LoRA
parameters, is frozen in this stage, while the memory encoder, memory
attention, stage-related parameters, and both action branches remain
trainable. This separates visual adaptation from learning to use interaction
history. Both stages use eight GPUs, mixed-precision training, and no
gradient accumulation.

\begin{table}[h]
    \centering
    \caption{\textbf{Training hyperparameters of SEEK on HIDE.}
    Stage~1 learns the memory-free policy; Stage~2 trains the memory-augmented
    policy from the Stage~1 checkpoint. Batch sizes are global across all GPUs.}
    \label{appendix_tab:hyperparam_seek_hide}
    \vspace{0.2em}
    % \resizebox{0.95\linewidth}{!}{%
    \begin{tabular}{@{}lcc@{}}
    \toprule
    Hyperparameter & Stage 1 & Stage 2 \\
    \midrule
    Number of GPUs                  & 8                    & 8 \\
    Batch size (frames)              & 56                   & 448 \\
    Batch size (temporal sequences)  & --                   & 64 \\
    Frames per sequence             & --                   & 7 \\
    Gradient accumulation           & 1                    & 1 \\
    Peak learning rate              & $7\times10^{-4}$     & $2\times10^{-4}$ \\
    Optimizer                       & LAMB + Adam$^{\dagger}$ & Adam \\
    Learning rate schedule          & Cosine decay         & Cosine decay \\
    Weight decay                    & $10^{-4}$            & 0 \\
    Warmup steps                    & 2,000                & 250 \\
    Checkpoint epoch$^{\ddagger}$   & 20                   & 10 \\
    Steps to checkpoint$^{\ddagger}$ & 57,140              & 3,575 \\
    Cosine horizon (steps)           & 114,280              & 7,150 \\
    LoRA rank                       & 16                   & 16 \\
    LoRA parameters                 & Trainable            & Frozen \\
    Coarse / fine action branches    & Trainable / trainable & Trainable / trainable \\
    Historical memory budget        & --                   & 6 entries \\
    Input resolution (per view)      & $224\times224$       & $224\times224$ \\
    \bottomrule
    \end{tabular}%
    % }
    \vspace{0.3em}
\begin{minipage}{0.95\linewidth}
\footnotesize
$^{\dagger}$Stage~1 uses LAMB for non-LoRA trainable parameters and Adam
updates for LoRA parameters. Its peak learning rate follows
$1.25\times10^{-5}\times B$, where $B=56$; Stage~2 uses a peak rate of
$2\times10^{-4}$ with 64 sequences of seven frames per batch.

$^{\ddagger}$The reported model uses the Stage~1 epoch-20 checkpoint and
Stage~2 epoch-10 checkpoint from schedules configured for 40 and 20 epochs,
respectively. The cosine horizons are retained rather than shortened to
the checkpoint epochs. Steps count scheduled updates, including any updates
skipped by mixed-precision loss scaling.
\end{minipage}
\end{table}

\subsection{Evaluation Metrics}
The primary metric is task success rate. On HIDE, we report per-task success rates, category averages over five tasks, and an overall unweighted average over all 15 tasks. Each reported model uses one checkpoint across all tasks. RLBench and \texttt{The Colosseum} follow their respective evaluation protocols, with clean and perturbed success rates reported separately for the latter.

\begin{figure}[htb]
    \centering
    \includegraphics[width=0.6\linewidth]{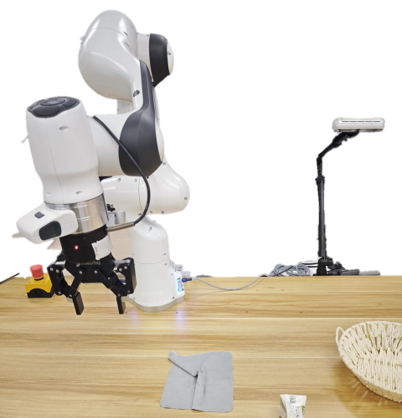}
    \caption{Scene of Real-world experiment.}
    \label{fig:real-scene}
\end{figure}

\subsection{Real-Robot Evaluation}
\label{app:real_world}
\begin{figure}[h]
    \centering

    \begin{subfigure}[t]{0.48\textwidth}
        \centering
        \includegraphics[width=\linewidth]{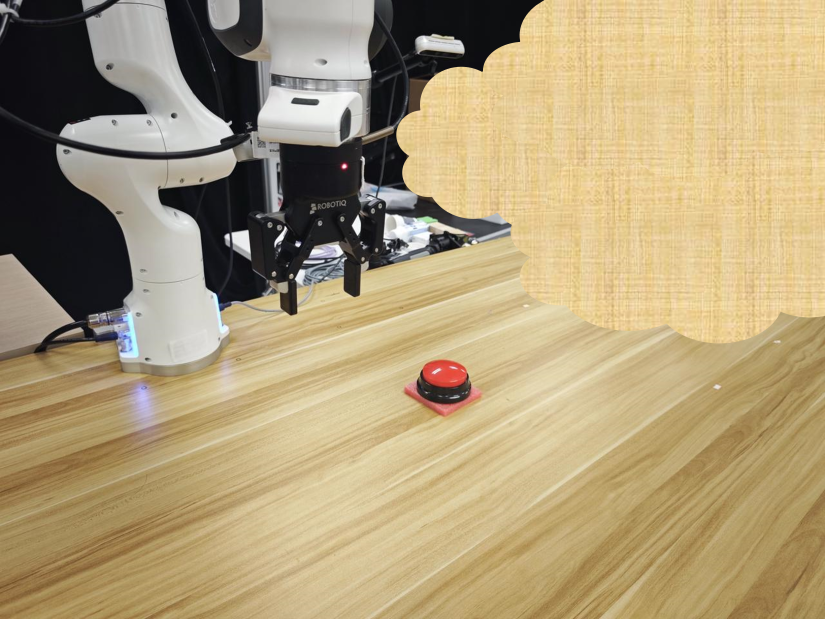}
        \caption{Repeated button pressing.}
        \label{fig:real_push}
    \end{subfigure}
    \hfill
    \begin{subfigure}[t]{0.48\textwidth}
        \centering
        \includegraphics[width=\linewidth]{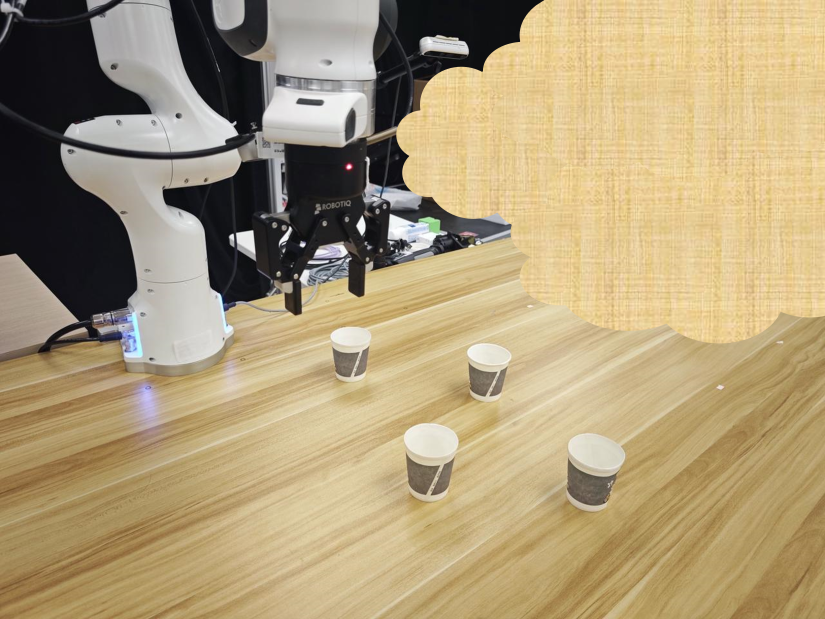}
        \caption{Cup stacking.}
        \label{fig:real_stack}
    \end{subfigure}

    \vspace{4pt}

    \begin{subfigure}[t]{0.48\textwidth}
        \centering
        \includegraphics[width=\linewidth]{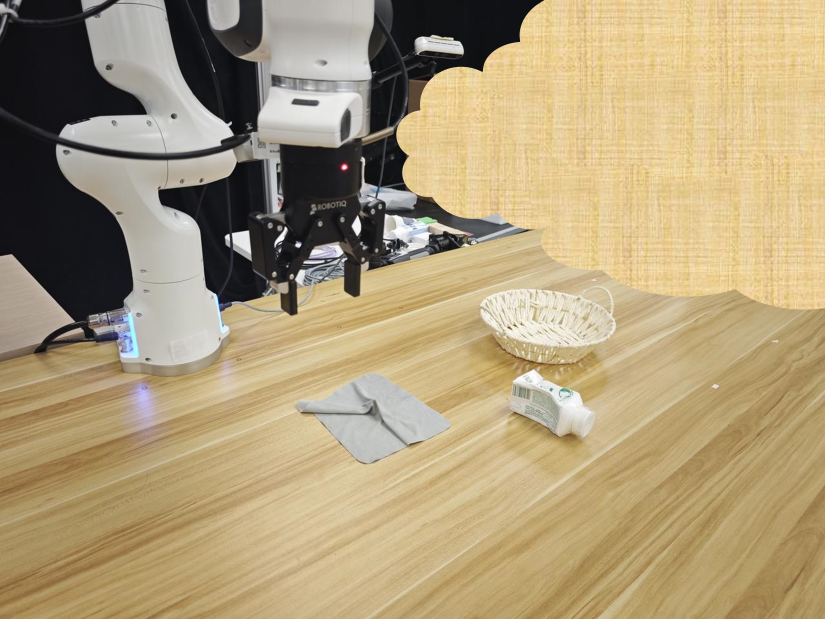}
        \caption{Desk cleaning.}
        \label{fig:real_clean}
    \end{subfigure}
    \hfill
    \begin{subfigure}[t]{0.48\textwidth}
        \centering
        \includegraphics[width=\linewidth]{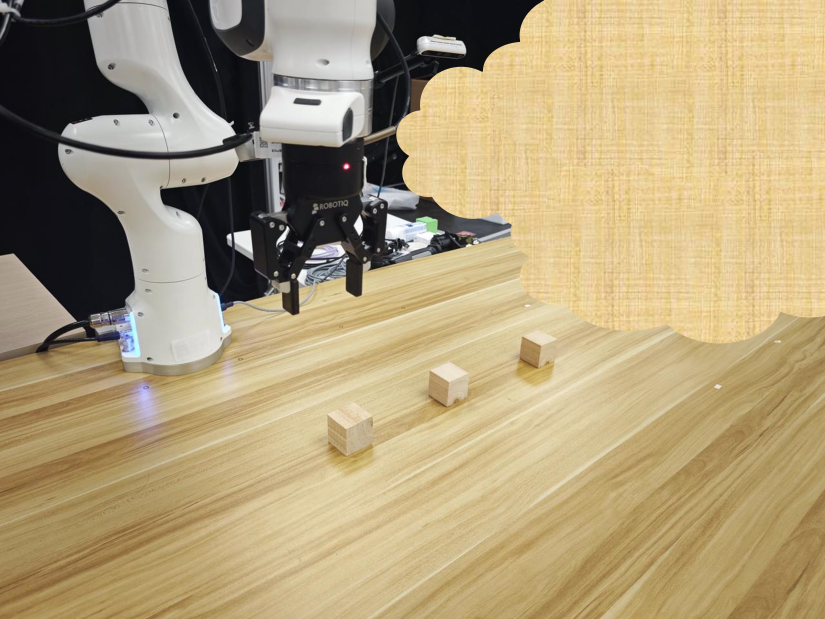}
        \caption{Block lifting and white-piece search.}
        \label{fig:real_lift}
    \end{subfigure}

    \caption{
        \textbf{Real-world task setups.}
        The four physical tasks used for real-robot evaluation:
        repeated button pressing, cup stacking, desk cleaning,
        and lifting blocks to search for a white piece.
    }
    \label{fig:real_world_tasks}
\end{figure}

We further evaluate on a Franka Emika Panda robot equipped with a
Robotiq gripper and an external Intel RealSense D455 RGB-D camera.
For each task, we collect 50 real-robot demonstrations and evaluate
the trained policy over 25 test trials, with the predefined task
variations uniformly represented during both data collection and
evaluation. The four tasks are pressing a button a specified number
of times, stacking a specified number of cups on the middle cup,
cleaning a desk, and lifting blocks to search for a white piece.
We compare SEEK with SAM2Act+~\citep{fang2025sam2act} and
$\pi_{0.5}$~\citep{physicalintelligence2025pi05}, and report success rates over the 25 evaluation trials.
All methods use separately trained real-robot policies. 
The scene and tasks are shown in
\cref{fig:real-scene,fig:real_world_tasks}.

%%%%%%%%%%%%%%%%%%%%%%%%%%%%%%%%%%%%%%%%%%%%%%%%%%%%%%%%%%%%%%%%%%%%%%%%%%%%%%%%%%%%%%% Results
\section{detailed results}
\label{app:detailed_results}
Here, we report the detailed results of SEEK and the baseline models on RLBench and The COLOSSEUM in~\cref{tab:rlbench,tab:colosseum}.
%%%%%%%%%%%%%%%%%%%%%%%%%%%%%%%%%%%%%%%%%%%%%%%%%%%%%%%%%%%%%%%%%%%%%%%%%%%%%% RLBench

\begin{table*}[h]
\centering
\caption{
\textbf{Full Comparisons of Multi-Task Performance on RLBench.}
We report success rates (\%) on 18 RLBench tasks \citep{james2020rlbench}.
SAM2Act results reported in the original paper are shown in light gray.
An asterisk ($^{*}$) denotes our evaluation using the officially released SAM2Act weights.
Their slightly lower performance under our unified evaluation protocol may reflect
differences in evaluation environments, dependency versions, and simulation stochasticity.
The last two rows report our method without and with memory, respectively.
Avg. Rank is recomputed across all listed rows by averaging per-task ranks,
with average ranks assigned to ties.
Bold and underlined values indicate the best and second-best results in each column
across all rows, respectively; ties are marked equally.
}
\label{tab:rlbench}

\setlength{\tabcolsep}{3pt}
\renewcommand{\arraystretch}{1.10}

\resizebox{\textwidth}{!}{
\begin{tabular}{l*{10}{c}}
\toprule
\textbf{Method}
& \shortstack{\textbf{Avg. Success}\\$\uparrow$}
& \shortstack{\textbf{Avg. Rank}\\$\downarrow$}
& \textbf{Close Jar}
& \textbf{Drag Stick}
& \textbf{Insert Peg}
& \shortstack{\textbf{Meat off}\\\textbf{Grill}}
& \shortstack{\textbf{Open}\\\textbf{Drawer}}
& \textbf{Place Cups}
& \textbf{Place Wine}
& \shortstack{\textbf{Push}\\\textbf{Buttons}} \\
\midrule

Image-BC (CNN) \citep{jang2022bcz}
& 1.3 & 14.08
& 0.0 & 0.0 & 0.0 & 0.0
& 4.0 & 0.0 & 0.0 & 0.0 \\

Image-BC (ViT) \citep{jang2022bcz}
& 1.3 & 14.25
& 0.0 & 0.0 & 0.0 & 0.0
& 0.0 & 0.0 & 0.0 & 0.0 \\

C2F-ARM-BC \citep{james2021c2f}
& 20.1 & 13.03
& 24.0 & 24.0 & 4.0 & 20.0
& 20.0 & 0.0 & 8.0 & 72.0 \\

HiveFormer \citep{guhur2022hiveformer}
& 45.3 & 11.06
& 52.0 & 76.0 & 0.0 & \textbf{100.0}
& 52.0 & 0.0 & 80.0 & 84.0 \\

PolarNet \citep{chen2023polarnet}
& 46.4 & 10.19
& 36.0 & 92.0 & 4.0 & \textbf{100.0}
& 84.0 & 0.0 & 40.0 & 96.0 \\

PerAct \citep{shridhar2023perceiver}
& 49.4 $\pm$ 4.3 & 9.83
& 55.2 $\pm$ 4.7 & 89.6 $\pm$ 4.1 & 5.6 $\pm$ 4.1 & 70.4 $\pm$ 2.0
& 88.0 $\pm$ 5.7 & 2.4 $\pm$ 3.2 & 44.8 $\pm$ 7.8 & 92.8 $\pm$ 3.0 \\

Act3D \citep{gervet2023act3d}
& 65.0 & 7.78
& 92.0 & 92.0 & 27.0 & 94.0
& \underline{93.0} & 3.0 & 80.0 & \underline{99.0} \\

RVT \citep{goyal2023rvt}
& 62.9 $\pm$ 3.7 & 8.11
& 52.0 $\pm$ 2.5 & \underline{99.2} $\pm$ 1.6 & 11.2 $\pm$ 3.0 & 88.0 $\pm$ 2.5
& 71.2 $\pm$ 6.9 & 4.0 $\pm$ 2.5 & 91.0 $\pm$ 5.2 & \textbf{100.0} $\pm$ 0.0 \\

RVT-2 \citep{goyal2024rvt}
& 81.4 $\pm$ 3.1 & 4.56
& \textbf{100.0} $\pm$ 0.0 & 99.0 $\pm$ 1.7 & 40.0 $\pm$ 0.0 & \underline{99.0} $\pm$ 1.7
& 74.0 $\pm$ 11.8 & 38.0 $\pm$ 4.5 & \underline{95.0} $\pm$ 3.3 & \textbf{100.0} $\pm$ 0.0 \\

3D Diffuser Actor \citep{ke20243dda}
& 81.3 & 4.81
& 96.0 $\pm$ 2.5 & \textbf{100.0} $\pm$ 0.0 & 65.6 $\pm$ 4.1 & 96.8 $\pm$ 1.6
& 89.6 $\pm$ 4.1 & 24.0 $\pm$ 7.6 & 93.6 $\pm$ 4.8 & 98.4 $\pm$ 2.0 \\

SAM-E \citep{zhang2024sam}
& 70.6 $\pm$ 0.7 & 6.22
& 82.4 $\pm$ 3.6 & \textbf{100.0} $\pm$ 0.0 & 18.4 $\pm$ 4.6 & 95.2 $\pm$ 3.3
& \textbf{95.2} $\pm$ 5.2 & 0.0 $\pm$ 0.0 & 94.4 $\pm$ 4.6 & \textbf{100.0} $\pm$ 0.0 \\

\samreported{SAM2Act \citep{fang2025sam2act}}
& \samreported{\textbf{86.8} $\pm$ 0.5} & \samreported{\underline{4.03}}
& \samreported{\underline{99.0} $\pm$ 2.0} & \samreported{99.0 $\pm$ 2.0} & \samreported{\underline{84.0} $\pm$ 5.7} & \samreported{98.0 $\pm$ 2.3}
& \samreported{83.0 $\pm$ 6.0} & \samreported{\textbf{47.0} $\pm$ 6.0} & \samreported{93.0 $\pm$ 3.8} & \samreported{\textbf{100.0} $\pm$ 0.0} \\

\midrule

SAM2Act$^{*}$ \citep{fang2025sam2act}
& 84.0 $\pm$ 1.0 & 4.14
& \textbf{100.0} $\pm$ 0.0 & \textbf{100.0} $\pm$ 0.0 & \textbf{92.0} $\pm$ 2.8 & \underline{99.0} $\pm$ 1.7
& 81.0 $\pm$ 3.3 & 25.0 $\pm$ 8.7 & 91.0 $\pm$ 3.3 & \textbf{100.0} $\pm$ 0.0 \\

\rowcolor{blue!6}
w/o memory (Ours)
& 84.1 $\pm$ 1.0 & 4.11
& \textbf{100.0} $\pm$ 0.0 & \textbf{100.0} $\pm$ 0.0 & 81.0 $\pm$ 8.2 & 98.0 $\pm$ 2.0
& 77.0 $\pm$ 5.9 & 37.0 $\pm$ 5.2 & \textbf{99.0} $\pm$ 1.7 & \textbf{100.0} $\pm$ 0.0 \\

\rowcolor{blue!6}
w/ memory (Ours)
& \underline{84.7} $\pm$ 0.6 & \textbf{3.81}
& \textbf{100.0} $\pm$ 0.0 & \textbf{100.0} $\pm$ 0.0 & 80.0 $\pm$ 8.9 & 98.0 $\pm$ 2.0
& 78.0 $\pm$ 4.5 & \underline{44.0} $\pm$ 7.5 & 94.0 $\pm$ 4.5 & \textbf{100.0} $\pm$ 0.0 \\

\midrule[0.8pt]
\textbf{Method}
& \shortstack{\textbf{Put in}\\\textbf{Cupboard}}
& \shortstack{\textbf{Put in}\\\textbf{Drawer}}
& \shortstack{\textbf{Put in}\\\textbf{Safe}}
& \textbf{Screw Bulb}
& \textbf{Slide Block}
& \textbf{Sort Shape}
& \shortstack{\textbf{Stack}\\\textbf{Blocks}}
& \textbf{Stack Cups}
& \shortstack{\textbf{Sweep to}\\\textbf{Dustpan}}
& \textbf{Turn Tap} \\
\midrule

Image-BC (CNN) \citep{jang2022bcz}
& 0.0 & 8.0 & 4.0 & 0.0
& 0.0 & 0.0 & 0.0
& 0.0 & 0.0 & 8.0 \\

Image-BC (ViT) \citep{jang2022bcz}
& 0.0 & 0.0 & 0.0 & 0.0
& 0.0 & 0.0 & 0.0
& 0.0 & 0.0 & 16.0 \\

C2F-ARM-BC \citep{james2021c2f}
& 0.0 & 4.0 & 12.0 & 8.0
& 16.0 & 8.0 & 0.0
& 0.0 & 0.0 & 68.0 \\

HiveFormer \citep{guhur2022hiveformer}
& 32.0 & 68.0 & 76.0 & 8.0
& 64.0 & 8.0 & 8.0
& 0.0 & 28.0 & 80.0 \\

PolarNet \citep{chen2023polarnet}
& 12.0 & 32.0 & 84.0 & 44.0
& 56.0 & 12.0 & 4.0
& 8.0 & 52.0 & 80.0 \\

PerAct \citep{shridhar2023perceiver}
& 28.0 $\pm$ 4.4 & 51.2 $\pm$ 4.7 & 84.0 $\pm$ 3.6 & 17.6 $\pm$ 2.0
& 74.0 $\pm$ 13.0 & 16.8 $\pm$ 4.7 & 26.4 $\pm$ 3.2
& 2.4 $\pm$ 2.0 & 52.0 $\pm$ 0.0 & 88.0 $\pm$ 4.4 \\

Act3D \citep{gervet2023act3d}
& 51.0 & 90.0 & 95.0 & 47.0
& 93.0 & 8.0 & 12.0
& 9.0 & 92.0 & 94.0 \\

RVT \citep{goyal2023rvt}
& 49.6 $\pm$ 3.2 & 88.0 $\pm$ 5.7 & 91.2 $\pm$ 3.0 & 48.0 $\pm$ 5.7
& 81.6 $\pm$ 5.4 & 36.0 $\pm$ 2.5 & 28.8 $\pm$ 3.9
& 26.4 $\pm$ 8.2 & 72.0 $\pm$ 0.0 & 93.6 $\pm$ 4.1 \\

RVT-2 \citep{goyal2024rvt}
& 66.0 $\pm$ 4.5 & 96.0 $\pm$ 0.0 & 96.0 $\pm$ 2.8 & 88.0 $\pm$ 4.9
& 92.0 $\pm$ 2.8 & 35.0 $\pm$ 7.1 & \textbf{80.0} $\pm$ 2.8
& 69.0 $\pm$ 5.9 & \textbf{100.0} $\pm$ 0.0 & 99.0 $\pm$ 1.7 \\

3D Diffuser Actor \citep{ke20243dda}
& \textbf{85.6} $\pm$ 4.1 & 96.0 $\pm$ 3.6 & \underline{97.6} $\pm$ 2.0 & 82.4 $\pm$ 2.0
& 97.6 $\pm$ 3.2 & 44.0 $\pm$ 4.4 & 68.3 $\pm$ 3.3
& 47.2 $\pm$ 8.5 & 84.0 $\pm$ 4.4 & \underline{99.2} $\pm$ 1.6 \\

SAM-E \citep{zhang2024sam}
& 64.0 $\pm$ 2.8 & 92.0 $\pm$ 5.7 & 95.2 $\pm$ 3.3 & 78.4 $\pm$ 3.6
& 95.2 $\pm$ 1.8 & 34.4 $\pm$ 6.1 & 26.4 $\pm$ 4.6
& 0.0 $\pm$ 0.0 & \textbf{100.0} $\pm$ 0.0 & \textbf{100.0} $\pm$ 0.0 \\

\samreported{SAM2Act \citep{fang2025sam2act}}
& \samreported{75.0 $\pm$ 3.8} & \samreported{\underline{99.0} $\pm$ 2.0} & \samreported{\textbf{98.0} $\pm$ 2.3} & \samreported{\underline{89.0} $\pm$ 2.0}
& \samreported{86.0 $\pm$ 4.0} & \samreported{64.0 $\pm$ 4.6} & \samreported{\underline{76.0} $\pm$ 8.6}
& \samreported{\underline{78.0} $\pm$ 4.0} & \samreported{\underline{99.0} $\pm$ 2.0} & \samreported{96.0 $\pm$ 5.7} \\

\midrule

SAM2Act$^{*}$ \citep{fang2025sam2act}
& 73.0 $\pm$ 7.1 & \underline{99.0} $\pm$ 1.7 & 92.0 $\pm$ 4.9 & \textbf{90.0} $\pm$ 4.5
& 85.0 $\pm$ 4.4 & \textbf{67.0} $\pm$ 3.3 & 39.0 $\pm$ 4.4
& \textbf{85.0} $\pm$ 5.2 & \textbf{100.0} $\pm$ 0.0 & 94.0 $\pm$ 4.5 \\

\rowcolor{blue!6}
w/o memory (Ours)
& 78.0 $\pm$ 2.0 & 98.0 $\pm$ 2.0 & 95.0 $\pm$ 4.4 & 83.0 $\pm$ 7.1
& 98.0 $\pm$ 2.0 & \underline{66.0} $\pm$ 4.5 & 42.0 $\pm$ 6.0
& 73.0 $\pm$ 5.2 & \textbf{100.0} $\pm$ 0.0 & 88.0 $\pm$ 5.7 \\

\rowcolor{blue!6}
w/ memory (Ours)
& \underline{80.0} $\pm$ 4.0 & \textbf{100.0} $\pm$ 0.0 & 97.0 $\pm$ 1.7 & 83.0 $\pm$ 3.3
& \textbf{100.0} $\pm$ 0.0 & 65.0 $\pm$ 3.3 & 40.0 $\pm$ 12.0
& \underline{78.0} $\pm$ 6.6 & \textbf{100.0} $\pm$ 0.0 & 87.0 $\pm$ 3.3 \\

\bottomrule
\end{tabular}
}
\end{table*}

%%%%%%%%%%%%%%%%%%%%%%%%%%%%%%%%%%%%%%%%%%%%%%%%%%%%%%%%%%%%%%%%%%%%%%%%%% colosseum
% Required packages:
% \usepackage{booktabs}
% \usepackage{graphicx}

\begin{table*}[t]
\centering
\caption{
\textbf{\texttt{The Colosseum} results.}
We compare the selected methods.
Entries report success rates (\%), with relative changes (\%) from the
corresponding unperturbed evaluation in parentheses.
Non-dagger literature results are aggregated over the perturbation-specific
task subsets used in SAM2Act. Relative changes use the Clean means
of the same task subsets.
Average is the unweighted mean over the 12 individual perturbations,
excluding All-Mixed; its relative change is measured against overall Clean.
PerAct and RVT are aggregated from the original benchmark paper;
RVT-2 is aggregated from the BridgeVLA paper;
SAM2Act uses paper-reported results.
Both memory variants are our methods.
$\dagger$ denotes legacy RVT results with different task subsets, retained
for reference and excluded from ranking.
The best and second-best displayed success rates among the listed
non-dagger methods are bold and underlined, respectively;
ties share the same rank.
}
\label{tab:colosseum}

\setlength{\tabcolsep}{4pt}
\renewcommand{\arraystretch}{1.15}

\resizebox{\textwidth}{!}{%
\begin{tabular}{l*{8}{c}}
\toprule
\textbf{Method}
& \textbf{Clean}
& \textbf{Average}
& \textbf{MO-Color}
& \textbf{RO-Color}
& \textbf{MO-Texture}
& \textbf{RO-Texture}
& \textbf{MO-Size}
& \textbf{RO-Size} \\
\midrule

PerAct
& $34.5\;(0.0)$
& $28.7\;(\downarrow 16.6)$
& $24.0\;(\downarrow 30.3)$
& $31.7\;(\downarrow 14.4)$
& $28.8\;(\downarrow 24.2)$
& $17.7\;(\downarrow 16.2)$
& $33.6\;(\downarrow 14.0)$
& $29.3\;(\downarrow 15.2)$ \\

RVT$^{\dagger}$
& $43.6\;(0.0)$
& $36.3\;(\downarrow 16.7)$
& $26.0\;(\downarrow 40.4)$
& $31.3\;(\downarrow 29.3)$
& $44.8\;(\uparrow 4.7)$
& $41.1\;(\uparrow 4.3)$
& $35.3\;(\downarrow 16.3)$
& $40.5\;(\downarrow 18.2)$ \\

RVT-2
& $67.8\;(0.0)$
& $59.5\;(\downarrow 12.3)$
& $53.0\;(\downarrow 21.8)$
& $59.2\;(\downarrow 9.8)$
& $\mathbf{59.7}\;(\downarrow 8.3)$
& $\underline{56.7}\;(\downarrow 0.3)$
& $60.9\;(\downarrow 5.9)$
& $53.4\;(\downarrow 17.2)$ \\

SAM2Act
& $64.7\;(0.0)$
& $\mathbf{62.3}\;(\downarrow 3.7)$
& $\mathbf{65.0}\;(\uparrow 0.4)$
& $61.0\;(\downarrow 0.9)$
& $\underline{55.4}\;(\downarrow 10.0)$
& $50.3\;(\downarrow 9.6)$
& $\mathbf{65.4}\;(\downarrow 7.5)$
& $\mathbf{58.0}\;(\downarrow 16.2)$ \\

\midrule

\rowcolor{blue!6}
w/o memory (Ours)
& $\underline{68.4}\;(0.0)$
& $61.5\;(\downarrow 10.1)$
& $62.1\;(\downarrow 9.3)$
& $\underline{65.2}\;(\downarrow 1.2)$
& $52.1\;(\downarrow 11.1)$
& $\mathbf{56.8}\;(\downarrow 12.6)$
& $\underline{63.1}\;(\downarrow 10.3)$
& $\underline{57.3}\;(\downarrow 20.6)$ \\

\rowcolor{blue!6}
w/ memory (Ours)
& $\mathbf{68.6}\;(0.0)$
& $\underline{61.9}\;(\downarrow 9.8)$
& $\underline{62.4}\;(\downarrow 9.0)$
& $\mathbf{65.7}\;(\downarrow 0.9)$
& $52.7\;(\downarrow 11.0)$
& $56.6\;(\downarrow 12.9)$
& $62.4\;(\downarrow 11.4)$
& $\underline{57.3}\;(\downarrow 20.6)$ \\

\midrule
\textbf{Method}
& \shortstack{\textbf{Light}\\\textbf{Color}}
& \shortstack{\textbf{Table}\\\textbf{Color}}
& \shortstack{\textbf{Table}\\\textbf{Texture}}
& \textbf{Distractor}
& \shortstack{\textbf{Background}\\\textbf{Texture}}
& \shortstack{\textbf{Camera}\\\textbf{Pose}}
& \multicolumn{2}{c}{\textbf{All-Mixed}} \\
\midrule

PerAct
& $29.1\;(\downarrow 15.5)$
& $30.4\;(\downarrow 11.9)$
& $23.2\;(\downarrow 32.7)$
& $27.1\;(\downarrow 21.3)$
& $33.5\;(\downarrow 2.8)$
& $36.3\;(\uparrow 5.4)$
& \multicolumn{2}{c}{$7.2\;(\downarrow 79.1)$} \\

RVT$^{\dagger}$
& $34.0\;(\downarrow 22.0)$
& $30.0\;(\downarrow 31.2)$
& $45.2\;(\uparrow 3.7)$
& $18.8\;(\downarrow 56.9)$
& $46.4\;(\uparrow 6.4)$
& $42.2\;(\downarrow 3.2)$
& \multicolumn{2}{c}{$6.4\;(\downarrow 85.3)$} \\

RVT-2
& $58.0\;(\downarrow 14.4)$
& $62.6\;(\downarrow 7.8)$
& $56.6\;(\downarrow 16.5)$
& $\underline{60.8}\;(\downarrow 10.5)$
& $\mathbf{68.7}\;(\uparrow 1.3)$
& $\underline{64.4}\;(\downarrow 5.0)$
& \multicolumn{2}{c}{$15.6\;(\downarrow 77.1)$} \\

SAM2Act
& $\underline{65.2}\;(\uparrow 0.7)$
& $65.6\;(\uparrow 1.3)$
& $\mathbf{65.4}\;(\uparrow 1.0)$
& $\mathbf{62.3}\;(\downarrow 3.8)$
& $\underline{68.6}\;(\uparrow 6.0)$
& $\mathbf{65.7}\;(\uparrow 1.6)$
& \multicolumn{2}{c}{$\mathbf{26.9}\;(\downarrow 58.4)$} \\

\midrule

\rowcolor{blue!6}
w/o memory (Ours)
& $\mathbf{65.5}\;(\downarrow 4.2)$
& $\underline{66.4}\;(\downarrow 2.9)$
& $60.3\;(\downarrow 11.8)$
& $58.4\;(\downarrow 14.6)$
& $67.3\;(\downarrow 1.6)$
& $63.2\;(\downarrow 7.6)$
& \multicolumn{2}{c}{$25.3\;(\downarrow 63.0)$} \\

\rowcolor{blue!6}
w/ memory (Ours)
& $64.9\;(\downarrow 5.4)$
& $\mathbf{69.9}\;(\uparrow 1.9)$
& $\underline{60.7}\;(\downarrow 11.5)$
& $58.6\;(\downarrow 14.6)$
& $67.6\;(\downarrow 1.5)$
& $63.6\;(\downarrow 7.3)$
& \multicolumn{2}{c}{$\underline{26.3}\;(\downarrow 61.7)$} \\

\bottomrule
\end{tabular}%
}
\end{table*}

\end{document}